\documentclass[journal,twoside,web]{ieeecolor}
\usepackage{generic}
\usepackage{amsmath,amssymb,amsfonts}
\usepackage{algorithmic}
\usepackage{graphicx}
\usepackage{algorithm,algorithmic}
\usepackage{hyperref}
\usepackage{threeparttable}
\usepackage[sorting=none,style=ieee]{biblatex}
\usepackage{booktabs}
\usepackage{multirow}

\usepackage{color,soul}
\usepackage[table,xcdraw]{xcolor}

\definecolor{myblue}{RGB}{0, 102, 204}
\usepackage{textcomp}
\begin{document}
\title{PosDiffAE: Position-aware Diffusion Auto-encoder For High-Resolution Brain Tissue Classification Incorporating Artifact Restoration  
}
\author{Ayantika Das, Moitreya Chaudhuri, Koushik Bhat, Keerthi Ram, Mihail Bota, Mohanasankar~ Sivaprakasam, \IEEEmembership{Member, IEEE} \thanks{Ayantika Das, Moitreya Chaudhuri, Koushik Bhat, and Mohanasankar Sivaprakasam  are with the Department of Electrical Engineering, Indian Institute of Technology Madras (IITM), Chennai
600036, India. Keerthi Ram, Mihail Bota, and Mohanasankar Sivaprakasam are with Sudha Gopalakrishnan Brain Centre (SGBC), IITM, Chennai 600036, India. Corresponding author: Ayantika Das (email: dasayantika486@gmail.com).}
}

\maketitle

\begin{abstract}

Denoising diffusion models produce high-fidelity image samples by capturing the image \textit{distribution} in a progressive manner while initializing with a simple distribution and compounding the distribution complexity.
Although these models have unlocked new applicabilities, the sampling mechanism of diffusion does not offer means to extract image-specific semantic representation, which is inherently provided by auto-encoders. The encoding component of auto-encoders enables mapping between a specific image and its latent space, thereby offering explicit means of enforcing structures in the latent space.
By integrating an encoder with the diffusion model, we establish an auto-encoding formulation, which learns image-specific representations and offers means to organize the latent space.

In this work, \textit{First,} we devise a mechanism to structure the latent space of a diffusion auto-encoding model, towards recognizing region-specific cellular patterns in brain images. We enforce the representations to regress positional information of the patches from high-resolution images. This creates a conducive latent space for differentiating tissue types of the brain. \textit{Second,} we devise an unsupervised tear artifact restoration technique based on neighborhood awareness, utilizing latent representations and the constrained generation capability of diffusion models during inference. \textit{Third,} through representational guidance and leveraging the inference time steerable noising and denoising capability of diffusion, we devise an unsupervised JPEG artifact restoration technique. Through experimentation, we validate that the representations of our model show (i) \textit{multi-tasking} capability (region classification, position regression, and artifact restoration), (ii) \textit{robustness} in artifact conditions, and (iii) \textit{generalization} across fetal and adult human brains. This demonstrates the effectiveness of representations of our model, PosDiffAE, enabling multiple unsupervised downstream tasks.

\end{abstract}


\section{Introduction}

\label{sect:intro}  % \label{} allows reference to this section
%%%%%%%%%%%%%%%%%%%%%%%%%%%%%%%%%%%%%%%%%%%%%%%

%%%%%%%%%%%%%%%%%%%%%%%%%%%%%%%%
Brain tissue processing using histological techniques, followed by High-resolution (HR) imaging, provides detailed cellular-level visual information of brain structures (brain regions, fiber tracts, and ventricles) and can be employed successfully in post-mortem investigations of brain architecture. One of the most employed tissue staining methods, Nissl staining for neuronal cell bodies, is used to highlight the cellular distribution and regional density variations, which are used to define and delineate regions and nuclei in the brain~\cite{WATSON2010153}, besides pinpointing the spatial location of individual brain cells. Digitizing serial sections of whole human brains at various developmental stages, from fetal onward \cite{ding2022cellular}, collates a timeline view of brain development. 
This can inform and enhance routine clinical imaging, interpretation, diagnosis, and staging of developmental neurological conditions, serving as reference snapshots for visualizing and comparing developing brains.

Histological imaging of a whole organ is an intensive exercise, made possible by high-throughput scanners that produce gigapixel images per section. To process such high-volume, high-velocity scanned images calls for computer vision and machine intelligence techniques, which can supplement human expert interpretation at scale. A predominant approach in recent times is the use of deep learning models for image understanding, exemplified by successful application in automated cancer grading \cite{coudray2018classification} and neuronal cell detection and analysis \cite{kiwitz2020deep} in the field of digital pathology.

\textbf{Visual features to be modeled} include the textural information and spatial location of brain regions. The data-related obstacles to be handled include photometric variations and tissue processing-related artifacts manifested in the image, like tissue tears and compression-related artifacts. Furthermore, the information in whole slide images is resolution-specific - e.g., tumor boundaries are well perceived in cancer at 10$\times$ magnification, but 40$\times$ magnification is necessary for the detection of mitotic figures of cells. This implies that larger receptive fields are generally required for region delineation. A standard approach to delineate regions would be to use multi-resolution analysis, going to lower resolutions from the original cellular resolution. \textbf{Our objective} is to extract the necessary and sufficient information from a patch of finite size \textbf{at the original resolution} and use it to identify the brain region from which the patch was selected.

Convolutional neural networks (CNN) architectures such as ResNet and VGG are a staple method for extracting visual features. Even though these backbone models were originally trained for image classification, the feature maps computed by these pre-trained models are considered suitable for various image recognition tasks. However, we seek an alternative that is devoid of irrelevant data biases from visual patterns in natural images. Specifically, \textbf{Our first aim is to develop a task-agnostic representation}, using an unsupervised Auto-encoding (AE) approach, which extracts a sparse representation of Nissl image patches at original resolution. Secondly, to achieve a specific organization of the latent space where the representations of patches from the same anatomical region are bounded by simple linear hyperplanes.
\\
\textbf{Unsupervised methods}: A recent method for learning image representations involves vision transformer models trained in a self-supervised manner using knowledge distillation with no labels (DINO) \cite{caron2021emerging}. Such training has been shown to organize the feature space in an interpretable structure, with similar images getting embedded proximally in the feature space. This result, while significant, is specific to the particular combination of DINO training and vision transformer architecture. 
\\
Variational Auto-encoders (VAE) \cite{kingma2013auto,wei2020recent} are a family of generative models that use an encoder model to map the distribution of input (image) data $x$ into a latent embedding distribution of chosen form and a decoder model to generate samples from the input distribution, given the latent embedding $z$. Training a VAE essentially forms a latent representation that captures sufficient information about the input data to maximize (a lower bound on) the probability of generating realistic samples from the input distribution.

Built upon the distributional capturing capabilities of VAE, \textbf{Denoising Diffusion Models} (DDMs) are new generative modeling approaches that can effectively learn the target distributions of images $p(x)$ and generate high-fidelity, realistic images. Furthermore, if the Gaussian priors ($\mathcal{N}(\mu, \sigma)$) are in the image domain ($\mu, \sigma \in \mathbb{R}^{3 \times h \times w}$) during the iterative noising and denoising process of DDM, then the encoded latent embeddings ($z$) of images are not restricted to specific distributional enforcement \cite{ho2020denoising}, \cite{nichol2021improved}. This allows the latent vectors to capture multiple modes available in the data distribution flexibly, unlike VAE or Generative Adversarial Networks (GANs), where the encoded latent embeddings are confined to lie near certain desired modes. 
 % \cite{ho2022cascaded}
Although the encoded latent vectors capture the data distribution, the unconditional formulation of DDMs does not embed properties of the specific image in the latent vector. In order to implement conditioning while inheriting all the latent representational capabilities of DDMs, \textbf{we propose to extract the latent vector} ($z$) from an encoding-based DDM, which is subsequently leveraged to classify image patches into different regions of the fetal and adult Nissl brain images. The encoding component is tasked with the extraction of image-specific representational features to guide the DDM process ($p(x|z)$) to generate the same (input) image, forming a DDM-based AE structure $DiffAE$ \cite{preechakul2022diffusion}. 

The $DiffAE$ model extracts meaningful features in a task-agnostic and unsupervised manner; this makes the modeling process unaware of the spatial positional information of the patches, which is lost during the process of the patch creation. In order to introduce positional awareness while operating at the original resolution, \textbf{we propose to incorporate the spatial information} of the patches in the latent representation of the model. This incorporation is similar to awareness used by annotators while demarcating different regions of the brain. Additionally, the positional information acts as a weak proxy, guiding the modeling process without being dependent on supervision with annotated labels.

The process of inference of DDM-based models allows image generation by steering through task-specific constraints, which enables the implicit capability to perform in-painting tasks in an unsupervised manner \cite{lugmayr2022repaint}. Although DDMs have been applied to restore artifacts in histological images \cite{he2023artifact, kropp2023denoising}, they do not ensure continuous restoration across patches in HR images. Therefore, \textbf{we propose an unsupervised tear artifact restoration} technique using DDMs by leveraging contextual information from neighboring patches. Additionally, the DDM-based models allow adaptive noising and denoising strategies to generate images during inference. \textbf{We propose} to employ these strategies to develop an unsupervised technique for blind \textbf{JPEG compression} artifact restoration. 
% Based on this, we propose an unsupervised technique for the recovery of JPEG-compressed images. 
Our contributions can be highlighted as follows:
\begin{itemize}
    \item We propose $PosDiffAE$, a position-aware DDM-based AE model that can extract meaningful representations of HR image patches while being trained in an unsupervised task-agnostic manner. We show that these representations can perform (i) discrimination of various anatomical regions of fetal and adult human brains and (ii) regression of spatial information of an image patch.
    
    \item Further leveraging upon the steerable inference-time image generation capability of DDMs, we propose unsupervised Artifact restoration techniques utilizing $PosDiffAE$ for ensuring robustness in handling histology-specific artifacts, tissue tears, and JPEG compression.

    \item We validate the region classification, position regression, and artifact restoration capability of our model in variable data setups and under various artifact settings. We have evaluated the region classification task in five human brains, constituting four fetal brains from four different gestational ages and one adult human brain of a $34$-year-old, from two different data acquisition sites, demonstrating the generalization ability of our model.
\end{itemize}

%%%%%%%%%%%%%%%%%%%%%%%%%%%%%%%%%%%%%
%%%%%%%%%%%%%%%%%%%%%%%%%%%%%%%%%%%%%
\begin{figure*}[t]
  \centering
  \includegraphics[width=1.05\linewidth]{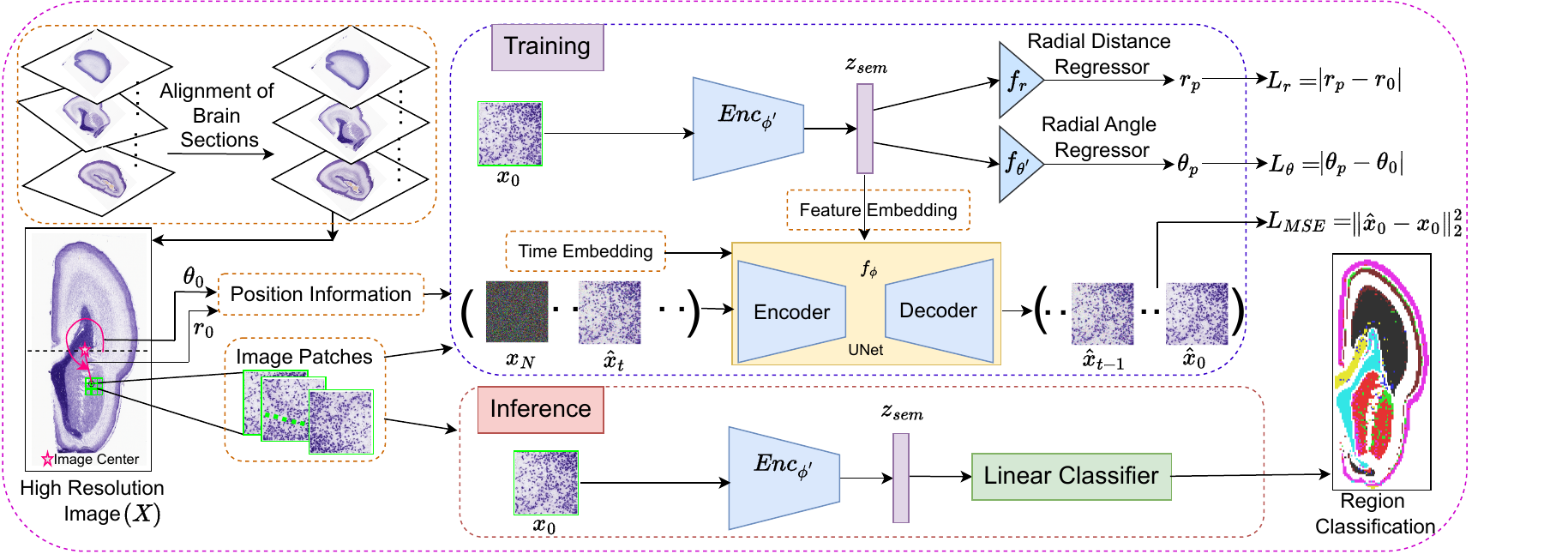}
\caption{From left to right, (i) \textit{Alignment}-  brain sections are aligned, (ii) \textit{Extraction}-  image patches and positional data (radial angles and distances) are extracted from aligned brain sections, (iii) \textit{Training Block}- patches and their positions are utilized for training the encoder ($Enc_{\phi^{'}}$) and denoising UNet ($f_{\phi}$), and (iv) \textit{Inference Block}- the encoded representations ($z_{sem}$) from the trained $Enc_{\phi^{'}}$ classify regions. In the (iii) \textit{Training Block}, the encoding of an image, regression of $z_{sem}$, the flow of denoising with $f_{\phi}$, and the losses associated are depicted.}
\label{fig:architechture}
\end{figure*}

%%%%%%%%%%%%%%%%%%%%%%%%%%%%%%%%%%%%%

\section{Related works}
\label{sect:related_work}
We categorize our method as unsupervised modeling of HR histology images, a denoising diffusion-based generative model, and an unsupervised artifact restoration model. We present related work in these three categories, highlighting the ones taken up for experimental comparison.
\subsection{Self-supervised Learning in HR Histological Images} 

In order to reduce reliance on human annotations, there has been a shift towards Self-Supervised Learning (SSL) methods in histological image analysis. The SSL methods learn representations of images to perform several downstream tasks. This learning of representations can be abstracted into three categories, as discussed below.

\textit{Pre-text Task:} For disease stratification in histopathological images, pre-text-based learning is adopted to extract image patch representations. These tasks include pathology-specific predictions, such as magnification level and stain channels, and pathology-agnostic tasks, such as geometric transformation prediction and Auto-encoding \cite{koohbanani2021self}, \cite{srinidhi2022self}, \cite{ding2023tailoring}.

\textit{Contrastive Learning:} With the integration of contrastive learning into SSL, there has been a notable shift in approach. This has also yielded better results in downstream discriminative tasks for histopathological image analysis in disease stratification \cite{yang2021self}, \cite{zhao2023self}, \cite{tan2023histopathology}, and region grouping in adult human brain cortex \cite{schiffer2021contrastive}.

\textit{Transformers:} With the advent of transformers, these networks have been widely applicable for representation learning in histological data analysis \cite{hu2023state}, \cite{huang2021integration}, \cite{wang2022transformer}. These networks have efficient attention mechanisms \cite{han2022survey}, which enable enhanced similarity extraction strategies, such as self-distillation \cite{caron2021emerging}. Leveraging self-distillation, a hierarchical patch-wise representation extractor for histopathological images (HIPT) was developed \cite{chen2022scaling}.
Utilizing the attention mechanism of transformers and similarity extraction methods, a kernel attention-based strategy was tailored for histological tasks (KAT) \cite{zheng2023kernel}.
However, all these methods, depending on patch-wise similarity maximization, often rely heavily on data augmentation. Masked Auto-encoder (MAE) \cite{he2022masked} introduced a new SSL paradigm, mitigating this dependency and also showing effectiveness in generating representations for histological image analysis \cite{luo2023self, horst2024cellvit}. Nevertheless, these transformer models are heavily dependent on large data sets, thus posing a constraint for medical image analysis \cite{dai2023swin}.
In order to alleviate this data dependency, the Swin-transformer-based MAE model (SwinMAE) was utilized \cite{ma2023efficient}.

\subsection{Denoising Diffusion Models (DDMs) }
DDMs have emerged as the state-of-the-art technique for generating high-fidelity images while processing natural images \cite{ho2020denoising}, \cite{nichol2021improved}. DDMs are also able to capture complex cell-level details in HR histopathological image generations \cite{moghadam2023morphology}, \cite{xu2023vit}, \cite{harb2024diffusion}. The effective generation capability is a consequence of the semantic-rich latent representations of the models. These latent representations have primarily been explored to perform attribute manipulation \cite{epstein2023diffusion, zhang2022unsupervised, preechakul2022diffusion}. The utilization of these representations for downstream discriminative tasks was done by employing different search-based techniques \cite{xiang2023denoising}, \cite{yang2023diffusion} to find the optimal time representation. However, we propose the usage of the representations from Diffusion AEs \cite{preechakul2022diffusion} for region discrimination in histological fetal brain images.

\subsection{DDM-based Unsupervised Artifact Restoration }
DDMs offer enhanced flexibility for generating images under various constraints \cite{nair2023unite}. Several unsupervised approaches have been presented for artifact restoration tasks \cite{corneanu2024latentpaint}, \cite{kawar2022denoising}, \cite{fei2023generative}, \cite{lugmayr2022repaint}. In the realm of artifact restoration in HR histological images, an unsupervised in-painting-based sampling \cite{lugmayr2022repaint} approach is utilized in ArtiFusion \cite{he2023artifact} and HARP \cite{fuchs2024harp}. 
However, these methods do not fully leverage the contextual information. We propose to incorporate context from neighboring image patches, which is essential while restoring HR histological images. For the restoration of JPEG compressed images using DDMs, inference-based non-blind (DDRM) \cite{kawar2022denoising} and blind techniques \cite{welker2024driftrec} have been employed. We adopt similar blind techniques and integrate contextual guidance to achieve JPEG restoration.

%%%%%%%%%%%%%%%%%%%%%%%%%%%%%%%%%%%%
\begin{figure*}[!hbt]
  \centering
  \includegraphics[width=1\linewidth]{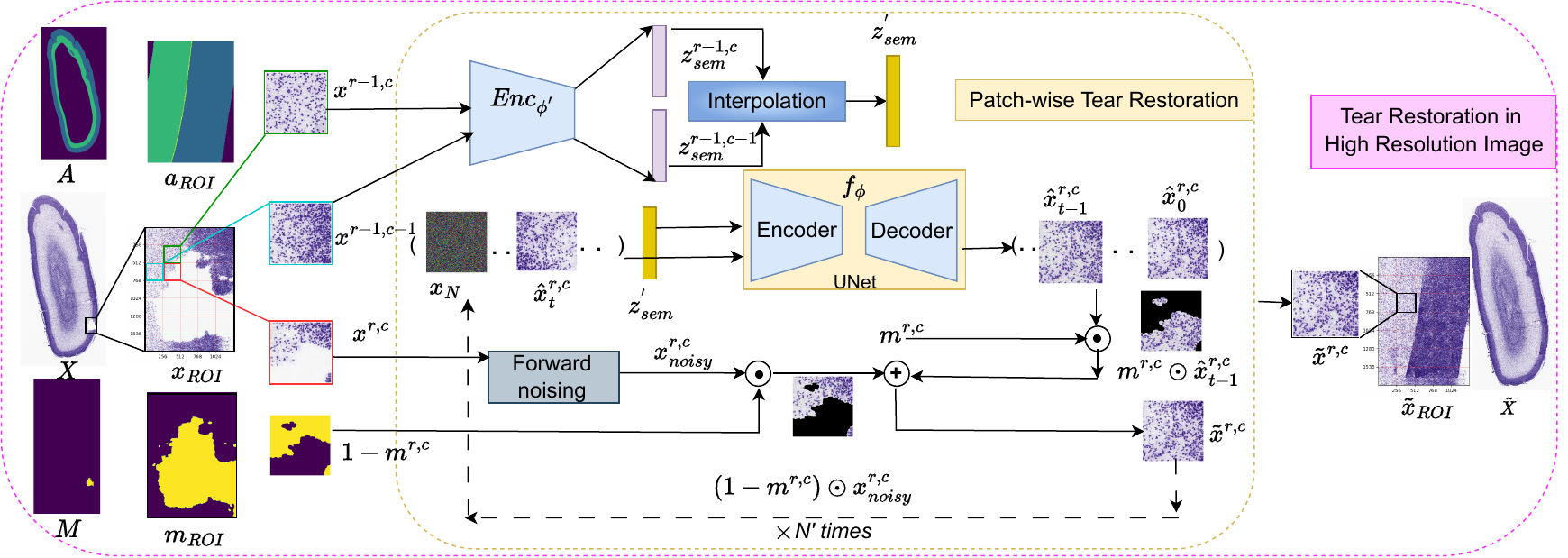}
\caption{From left to right, (i) \textit{High-resolution and ROI}- the first and second columns have annotations, artifact images, and masks for HR and ROI, (ii) \textit{Patch inputs}- the third column has artifact patch, neighborhood clean patches and artifact mask patches, (iii) \textit{Patch-wise restoration block}- artifact patch with the neighborhood are processed through the encoding mechanism along with the mask constrained denoising, (iv) \textit{Restored outputs}- restored patch, ROI image, and high-resolution image are depicted.}

\label{fig:Tear_Artifact_comb}
\end{figure*}

%%%%%%%%%%%%%%%%%%%%%%%%%%%%%%%%%%%%

\section{Methodology}
\label{sect:method}

 $PosDiffAE$ learns a specific mapping between an image and the latent space, producing an encoded image representation ($z_{sem}$), integrable into the conventional denoising diffusion structure and capable of guiding the iterative generation of image samples from noise. The $z_{sem}$ of $PosDiffAE$ is processed to regress positional details of image patches, which helps structure the latent space in a more region-specific manner. We develop a brain region $classifier$ by finding suitable hyperplanes within this space. The task of discriminating regions and assigning region-wise labels to patches from HR images can be formulated as $a_{reg} = classifier(x_{0})$, where $x_{0} \in \mathbb{R}^{3 \times h \times w}$ represents a patch from HR image ($ X \in \mathbb{R}^{3 \times H \times W} $ ) and $a_{reg}$ represents the region label assigned to $x_{0}$. The region labels $a_{reg}$ belong to the set of regions demarcated in ground truth annotation ($ A \in \mathbb{R}^{3 \times H \times W} $). Our proposed model is schematically detailed in Figure \ref{fig:architechture}. The details of the formulation of $DiffAE$ are elaborated in Subsection \ref{sec:DiffAE}.
 
 % The conventional DDM models use an attention-based UNet structure, which operates in a time-dependent manner. These consequently make the latent space of the UNet time-dependent, generating multiple latent variables and thus posing challenges in extracting a useful representation for a given input image \cite{preechakul2022diffusion}.  

\subsection{DDM-based Auto-encoding Network (DiffAE)}

\label{sec:DiffAE}
The $DiffAE$ structure integrates an additional learnable encoder ($Enc_{\phi^{'}}$), along with the conventional UNet structure ($f_{\phi}$) \cite{preechakul2022diffusion} of diffusion, to encode semantic information and offer a latent space. 
The latent space embedding generated by $Enc_{\phi^{'}}$ is given as, $z_{sem} = Enc_{\phi^{'}}(x_{0})$, where $z_{sem} \in \mathbb{R}^{f_{dim}}$.

The $DiffAE$ learns to conditionally map an arbitrary Gaussian Noise ($x_{T} \in \mathbb{R}^{3 \times h \times w}$) from the standard Gaussian distribution $\mathcal{N}(0,I)$ to an image sample ($x_{0}$), through iterative denoising process over time-steps $T$. At any time instant $t$, the goal of $DiffAE$ is to model a distribution $p_{\phi}(x_{t-1}|x_{t},z_{sem})$ in order to match the conditional prior distribution $q(x_{t-1}|x_{t},x_{0})$. The prior distribution is a Gaussian diffusion process and is represented as: 
\begin{equation}
    \label{eq:q_equation_1}
    q(x_t|x_{t-1}) = \mathcal{N}(\sqrt{1 - \beta_t} x_{t-1}, \beta_t I) 
\end{equation} 
where $ \beta_t $ are hyperparameters representing the noise levels. With Gaussian diffusion process, the noisy version of image $ x_0 $ at time $ t $ can be given as: 
\begin{equation}
    \label{eq:q_equation_2}
   q(x_t|x_0) = \mathcal{N}(\sqrt{\alpha_t} x_0, (1 - \alpha_t)I)  
\end{equation} 
where $ \alpha_t = \prod_{s=1}^{t} (1 - \beta_s) $ and $x_t = \sqrt{\alpha_t } x_0 + \sqrt{1 - \alpha_t} \epsilon_t$, $\epsilon_t \in \mathbb{R}^{3 \times h \times w}$.
Using \eqref{eq:q_equation_1} and \eqref{eq:q_equation_2}, the following form of the conditional prior distribution is derivable:
\begin{equation}
    \label{eq:q_equation_3}
    q(x_{t-1} | x_t, x_0) = \mathcal{N} (\sqrt{\alpha_{t-1}} x_0  + \sqrt{1 - \alpha_{t-1}} (\frac{x_t - \sqrt{\alpha_t} x_0 }  {\sqrt{1 - \alpha_t}}) ,0 ) 
    \end{equation} 
The distribution being modeled is given as follows:
\begin{equation}
\label{eq:p_equation_1}    
p_\phi ({x}_{0: T} | z_{{sem}})=p({x}_T) \prod_{t=1}^T p_\phi({x}_{t-1} | {x}_t, {z}_{{sem}})
\end{equation}

\begin{equation}
    \label{eq:p_equation_2}
    \small
p_\phi\left({x}_{t-1} \mid  {x}_t,  {z}_{{sem}}\right)= \begin{cases}\mathcal{N}\left( {f}_\phi\left( {x}_1, 1,  {z}_{\mathrm{sem}}\right),  {0}\right) & \text { if } t=1 \\ q\left( {x}_{t-1} \mid  {x}_t,  {f}_\phi\left( {x}_t, t,  {z}_{{sem}}\right)\right) & \text { otherwise }\end{cases}
\end{equation}
\\
where ${f}_\phi( {x}_t, t,  {z}_{{sem}})$ is parameterized as the output of the UNet model of $DiffAE$. The model is configured to directly estimate $x_0$ such that, $ \hat{x}_{0} = {f}_\phi( {x}_t, t,  {z}_{{sem}})$. 
This formulation, when optimized with $L_{{MSE}} = \| \hat{x}_{0} - x_0 \|_{2}^{2}$, produces a semantically rich representation ($z_{sem}$), as this representation explicitly guides the image formation.

We propose to incorporate positional information into this semantically rich $z_{sem}$ of $DiffAE$, by supervising them to regress the spatial details associated with a patch. The spatial location of a patch extracted from the HR image is expressed in terms of polar coordinates. This includes specifying the distance from the center of the image, along with the angle measured from the horizontal axis of the image. The details of $PosDiffAE$ are further elaborated in Subsection \ref{sec:PosDiffAE}.

\subsection{Proposed Network (PosDiffAE)} 

\label{sec:PosDiffAE}
% Positional regression components are incorporated in the architectural design, which enforces the latent embedding (${z}_{sem}$) to learn the positional information of the patches. The 
$PosDiffAE$ regresses (i) the radial distance ($r_{0}$) from the center ($c_{x},c_{y}$) of the HR image to the top-left corner of the patches ($x_{p},y_{p}$) and (ii) the radial angle $(\theta_{0}$) between $r_{0}$ and the horizontal axis passing through the center. The radial distance regressor and the radial angle regressor are parameterized with Fully connected linear layers, given as $r_{p} = f_{r}(z_{sem})$, $\theta_{p} = f_{\theta^{'}}(z_{sem})$. The optimization function with the additional constraints is given as $
L_{total} = \lambda_{1} L_{{MSE}} + \lambda_{2}  L_{r} + \lambda_{3} L_{\theta}$, where $ L_{r} = | {r}_{p} - {r}_{0} | $, $ L_{\theta} = |{\theta}_{p} - {\theta}_{0} |$, $\lambda_{1}$, $\lambda_{2}$ and $\lambda_{3}$ are hyper-parameters of the loss functions.

\subsubsection{Estimation of the Radial Distance and the Radial Angle}
% The Radial Distance ($r_{0}$) is a relative score from the centriod ($c_{x},c_{y}$) of $X_{WSI}$. 
% The centroid is evaluated using the background and foreground differentiation information in the HR image ($X$). The average positions of the pixel coordinates in the foreground of $X$ are weighted by their intensity values to estimate $c_{x}$ and $c_{y}$. Then, the euclidean distance between $c_{x}$, $c_{y}$ and $x_{p}$ and $y_{p}$ gives the resultant distance $r_{0}$. The $r_{0}$ value further undergoes a normalization process. More specifically the estimation of $r_{0}$ is given by, 
% $r_{0} = ((c_x - x_p)^2 + (c_y - y_p)^2)^{\frac{1}{2}}$. 

The centroid ($c_{x}$, $c_{y}$) of the HR image ($X$) is evaluated using the background and foreground differentiation information by calculating the average of the pixel coordinates, which lie within the foreground. Then, the Euclidean distance between $c_{x}$, $c_{y}$ and $x_{p}$, $y_{p}$ gives the resultant distance $r_{0}$. More specifically the estimation of $r_{0}$ is given by, 
$r_{0} = ((c_x - x_p)^2 + (c_y - y_p)^2)^{\frac{1}{2}}$. 
The obtained $r_{0}$ is further normalized by the following, $r_{0}  = \frac{r_{0}}{r_{X}}$, where $r_{X}$ is the maximum value of $r_{0}$ for the high resolution image ($X$) where the patch was extracted from.
% by the maximum value of $r_{0}$ from all the foreground patches of $X$. 
% (ii) the evaluation of relative angle ($\gamma$) for image origin coordinate alignment with the centriod 

For the radial angle ($\theta_{0}$) estimation, the following steps are required: (i) the evaluation of the angle ($\theta$) between $r_{0}$ and the horizontal line passing through centroid ($c_x$, $c_y$), and (ii) the evaluation of the alignment (rotation) angle ($\alpha$) between the different brain sections. The equation for estimation of angle $\theta$ is given as, $\theta = \arctan\left(\frac{c_y - y_p}{c_x - x_p}\right)$.

To align misaligned high-resolution brain section images, inter-section registration is performed using Scale-Invariant Feature Transform (SIFT) for feature extraction, Random Sample Consensus (RANSAC) for feature matching and rigid body transformation to estimate the inter-slice rotation angle ($\alpha$), with the overall radial angle is given by $\theta_{\text{0}} = \theta + \alpha $.

During inference of $PosDiffAE$, whenever the latent embedding ($z_{sem}$) is required, the image is given to $Enc_{\phi^{'}}(x_{0})$ and the encoded embedding is extracted. For image generation, we used the time-dependent sampling strategy of diffusion \cite{song2020denoising, preechakul2022diffusion} with the integration of $z_{sem}$. The generation is initiated with an arbitrary Gaussian noise. At any time instant $t$, we evaluate  $ \hat{x}_{0} = {f}_\phi( {x}_t, t,  {z}_{{sem}})$, using this predicted $\hat{x}_{0}$ we estimate $\hat{x}_{t-1}$ using Equation \ref{eq:q_equation_3}. This denoising process is done iteratively for $N$ number of time steps.

We utilize the inferencing mode of $PosDiffAE$ to propose a tear artifact restoration ($PosDiffAE+Restore$) technique that in-paints the tear-affected regions in the images. The inference time in-painting is achieved using the constrained image generation strategy \cite{lugmayr2022repaint} of diffusion models. This generation strategy is guided by the latent embeddings ($z_{sem}$) from $PosDiffAE$ to make the restoration process aware of the context of the neighboring patches. The constrained in-painting generation and the latent embedding-based guidance ensure patch-level and region-level consistency, respectively. This restoration strategy is detailed in Subsection \ref{sec:Tear_restore}.
% The restoration is performed in a patch-wise manner in the selected region of interest in the high resolution image.

\subsection{PosDiffAE+Restore: Tear Artifact}
\label{sec:Tear_restore}

 The schematic diagram for the tear artifact restoration pipeline is detailed in Figure \ref{fig:Tear_Artifact_comb}. Considering an HR brain section ($X$) and its corresponding ground truth annotation ($A$), an intensity threshold-based approach is used to detect the artifact regions, resulting in an artifact mask ($M \in \mathbb{R}^{3 \times H \times W}$). The restoration of $X$ is performed in regions where $M$ has non-zero pixels. The region of interest (ROI) is extracted from areas of $X$ where $M$ has non-zero pixels, forming image $x_{ROI}$. The corresponding ground truth annotation and the artifact mask for the extracted region are $a_{ROI}$ and $m_{ROI}$, respectively. A non-overlapping sliding window-based restoration approach is used within $x_{ROI}$, wherein each window size is kept compatible with the input size of $PosDiffAE$. 

The restoration pipeline at the patch level is detailed inside the yellow box in Figure \ref{fig:Tear_Artifact_comb}. Considering a certain row ($r$) and column ($c$) within $x_{ROI}$, we have a $h \times w$ window ($x^{r,c}$) and its corresponding mask ($m^{r,c}$). Our devised restoration technique ensures consistency at both the region and patch levels, respectively. In order to ensure region-level consistency, an interpolated conditioning vector is used to guide the generation process of the $PosDiffAE$ to be aware of the neighborhood. The interpolation of the conditioning vector is done from the neighboring patch windows following the equation below:
\begin{equation}
    \label{intrp_zsem}
    z^{'}_{sem} = (0.5)(Enc_{\phi^{'}}(x^{r-1,c}) + Enc_{\phi^{'}}(x^{r-1,c-1}))
\end{equation} 
where $x^{r-1,c}$ and $x^{r-1,c-1}$ refer to the patch on top and top left of the current patch $x^{r,c}$. Additionally, to enforce patch-level consistency, the mask $m^{r,c}$ of the patch is used to guide \cite{he2023artifact}, \cite{lugmayr2022repaint}, \cite{kropp2023denoising} the sampling process of $PosDiffAE$. For any time instant $t$, the predicted image is given as $\hat{x}^{r,c}_{0} = f_{\phi}(\Tilde{x}^{r,c}_{t}, {t} ,z^{'}_{sem})$. Obtaining the predicted image $\hat{x}^{r,c}_{0}$ and using it in Equation \ref{eq:q_equation_3}, the image ($\hat{x}^{r,c}_{t-1}$) for time instant $t-1$ can generated. The incorporation of the mask is done using the following equation:

\begin{equation}
    \label{mask_guidance}
    \Tilde{x}^{r,c}_{t-1} = (1 - m^{r,c}) \odot x^{r,c}_{noisy} + (m^{r,c}) \odot \hat{x}^{r,c}_{t-1}
\end{equation}
where $\odot$ represents the element-wise product, $x^{r,c}_{noisy}$ denotes the noisy version of $x^{r,c}$ which is sampled from the Gaussian distribution $\mathcal{N}(\sqrt{\alpha_{t}} x^{r,c}, (1 - \alpha_{t})I)$, $\Tilde{x}^{r,c}_{t-1}$ is the intermediate generation. This process is carried out for $N^{'}$ number of time-steps to generate the final restored image $\Tilde{x}^{r,c}$. These restored patches are stitched to generate artifact-free ROI image $\Tilde{x}^{ROI}$, generating the restored HR image $\Tilde{X}$.
%%%%%%%%%%%%%%%%%%%%%

We utilize the inferencing mode of $PosDiffAE$ to propose a JPEG restoration ($PosDiffAE+Restore$) technique that leverages the capability of diffusion models to generate images using an adaptive noising and denoising strategy. The level of noising and denoising parameter is chosen depending on the image compression quality. The latent embeddings ($z_{sem}$) of $PosDiffAE$ guide the restoration process by giving semantically meaningful information. This restoration strategy is further detailed in Subsection \ref{sec:jpeg_Compression}.

\subsection{PosDiffAE+Restore: JPEG Compression}
\label{sec:jpeg_Compression}
The image generation process of DDM models is an iterative denoising process, which is initiated from pure Gaussian Noise. These models provide flexibility by allowing the process to start from images with any level of added noise \cite{meng2021sdedit}. In such flexible noise addition cases, the number of denoising steps should be adaptively chosen based on the level of noise present in the image. The noising process resembles the effect in JPEG compression, initially affecting the high-frequency content in the images. Consequently, after gradually adding noise to the JPEG corrupted image, the information content within the noisy corrupted image resembles the noise-added clean image. The result is a reduction in the distributional disparity between noisy compressed and noisy clean images relative to their non-noisy counterparts. This strategy makes the distribution of the noisy compressed images tractable, thus allowing the sampling of clean images and recovery of the compressed images \cite{welker2024driftrec}.

However, this noising strategy involves a trade-off between realism and image faithfulness \cite{meng2021sdedit}. An increased level of noise in the image can mitigate artifacts due to compression, but it may compromise the fidelity of image generation. To address this challenge, the faithfulness of the generated images is ensured by the $z_{sem}$, while the noising strategy is tailored to prioritize realistic generations.

Considering $x_{compr}$ as the compressed image and using Equation \ref{eq:q_equation_2} the noisy compressed image ($x_{T^{'}}^{compr}$) is given as:
\begin{equation}
    x_{T^{'}}^{compr} = \sqrt{\alpha_{T^{'}} } x_{compr} + \sqrt{1 - \alpha_{T^{'}}} \epsilon_{T^{'}}
\end{equation}
where $T^{'}$ is the noising time step, which is adaptively chosen depending on the Quality Factor (QF) of the JPEG compressed image. Initiating from the noisy image $x_{T^{'}}^{compr}$ and integrating the vector $z^{compr}_{sem} = Enc_{\phi^{'}}(x_{compr})$, the model retrieves the non-compressed image through denoising utilizing adaptive time steps ($N^{''}$). 

\textit{Gamma Adjustment:} In order to check the robustness of our propositions, we have created a realistic data manipulation setup by introducing gamma adjustments ($x^\gamma$). It is imperative to assess the method with such adjustments since gamma correction is often used in the visualization of stained tissue.

%%%%%%%%%%%%%%%%%%%%%%%%%%%%%%%%%%%%%%
\begin{figure}[!t]
  \centering
  \centerline{\includegraphics[width=1\linewidth]{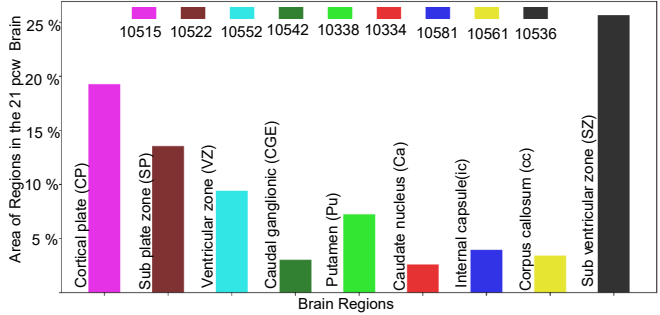}}
\caption{Area covered by different regions in the Allen $21$ pcw brain.}
\label{fig:Regions_present}
\end{figure}
%%%%%%%%%%%%%%%%%%%%%%%%%%%%%%%%%%%%%
%%%%%%%%%%%%%%%%%%%%%%%%%%%%%%%%%%%%%%%%%%%%%%%%%%%%%%%%%%%%%%%%%%%%%%%%%%%
\begin{table}[!tb]
\hfill
\centering
\caption{Parameter Details}
\begin{threeparttable}
\resizebox{0.9\columnwidth}{!}{%
\begin{tabular}{@{}l|l|l|l@{}}
\toprule
Parameters &
  Values &
  Parameters &
  Values \\ \midrule
Epochs &
  120 &
  \begin{tabular}[c]{@{}l@{}}$\lambda_{1}$, $\lambda_{2}$,\\ $\lambda_{3}$\end{tabular} &
  \begin{tabular}[c]{@{}l@{}}1, 0.001,\\ 0.001\end{tabular} \\ \midrule
% $h$, $w$ &
%   $256$, $256$ &
  \begin{tabular}[c]{@{}l@{}}h, w \\ H, W\end{tabular} &
  \begin{tabular}[c]{@{}l@{}}256, 256 pixels \\ Order of $1\mathrm{e}^{4}$, $1\mathrm{e}^{4}$ pixels \end{tabular} &
  \begin{tabular}[c]{@{}l@{}}$\theta_{0},\theta$ \\ $\alpha$\end{tabular} &
  \begin{tabular}[c]{@{}l@{}}$0-360^\circ$\\ $0-10^\circ$\end{tabular}
  \\ \midrule
Optimizer &
  Adam &
  $f_{dim}$ & $512$  \\ \midrule
Learning Rate &
  0.001 &
  $r_{0}$  &
  \begin{tabular}[c]{@{}l@{}}Depends on \\ H \& W\end{tabular} \\ \midrule
$f_{\phi}$, $Enc_{\phi^{'}}$ &
  \begin{tabular}[c]{@{}l@{}}Architecture \\ from DiffAE\end{tabular} &
  $N^{'}$ &
  $50$ \\ \midrule
\begin{tabular}[c]{@{}l@{}}$Classifier$ \\ $f_{r}$, $f_{\theta^{'}}$\end{tabular}    &
 \begin{tabular}[c]{@{}l@{}}Linear\tnote{1} \\ MLP Layers\end{tabular} &
  $T^{'}$ &
  \begin{tabular}[c]{@{}l@{}}20 (QF 5)\\ 20 (QF 10)\\ 10 (QF 15)\end{tabular} \\ \midrule
\begin{tabular}[c]{@{}l@{}}Diffusion Noise \\ ($\beta_{t}, T$)\end{tabular} &
  \begin{tabular}[c]{@{}l@{}}Linear Scheduling, 1000\end{tabular} &
  $N^{''}$ &
  \begin{tabular}[c]{@{}l@{}}50 (QF 5)\\ 40 (QF 10)\\ 40 (QF 15)\end{tabular} \\ \bottomrule
\end{tabular}%

}

\begin{tablenotes}
\item[1] \href{https://scikit-learn.org/stable/modules/generated/sklearn.svm.SVC.html}{scikit-learn SVM (SVC)}
\end{tablenotes}
\end{threeparttable}
\label{table:parameter_details}
\end{table}
%%%%%%%%%%%%%%%%%%%%%%%%
%%%%%%%%%%%%%%%%%%%%%%%%%%%%%%%%%%%%%%%%%%%%%%%%%%%%% TABLE

%%%%%%%%

% \usepackage{graphicx}
\renewcommand{\arraystretch}{0.8} 
\begin{table*}[!hbt]
\caption{The classification performance on the $21$ pcw validation dataset, reporting class-wise and overall metrics with Precision, Recall, and Accuracy. We provided parameter details and statistical analysis. }
% class-wise performance in terms of Precision/ Recall and overall performance with Accuracy.
% \usepackage{booktabs}
% \usepackage{graphicx}
% \begin{table}[]
\centering
\resizebox{1.7\columnwidth}{!}{%
\begin{tabular}{@{}ccccccccccccccc@{}}
\toprule
\begin{tabular}[c]{@{}c@{}}Region Numbers/\\ Names\end{tabular} &
  \begin{tabular}[c]{@{}c@{}}10334/\\ Ca\end{tabular} &
  \begin{tabular}[c]{@{}c@{}}10338/\\ Pu\end{tabular} &
  \begin{tabular}[c]{@{}c@{}}10515/\\ CP\end{tabular} &
  \begin{tabular}[c]{@{}c@{}}10522/\\ SP\end{tabular} &
  \begin{tabular}[c]{@{}c@{}}10536/\\ SZ\end{tabular} &
  \begin{tabular}[c]{@{}c@{}}10542/\\ VZ\end{tabular} &
  \begin{tabular}[c]{@{}c@{}}10552/\\ CGE\end{tabular} &
  \begin{tabular}[c]{@{}c@{}}10561/\\ cc\end{tabular} &
  \begin{tabular}[c]{@{}c@{}}10581/\\ ic\end{tabular} &
   &
  \multicolumn{2
  }{c}{\begin{tabular}[c]{@{}c@{}}Parameters\\ (M)\end{tabular}} \\ \midrule
Methods &
  \multicolumn{9}{c}{Precision ($\uparrow$)/ Recall ($\uparrow$)} &
  \begin{tabular}[c]{@{}c@{}}Accuracy \\ (\%)($\uparrow$)\end{tabular} &
  Encoder &
  \begin{tabular}[c]{@{}c@{}}Encoder $+$ \\ Decoder\end{tabular} &
  \begin{tabular}[c]{@{}c@{}}Time $/$ \\ Image \\ (s)\end{tabular} &
  \begin{tabular}[c]{@{}c@{}}Cohen's \\ Kappa \\ ($\kappa$) ($\uparrow$)\end{tabular} \\ \midrule
Trans-UNet\cite{chen2021transunet}  &
  \begin{tabular}[c]{@{}c@{}}0.48/\\ 0.57\end{tabular} &
  \begin{tabular}[c]{@{}c@{}}0.42/\\ 0.40\end{tabular} &
  \begin{tabular}[c]{@{}c@{}}0.53/\\ 0.63\end{tabular} &
  \begin{tabular}[c]{@{}c@{}}0.61/\\ 0.65\end{tabular} &
  \begin{tabular}[c]{@{}c@{}}0.51/\\ 0.37\end{tabular} &
  \begin{tabular}[c]{@{}c@{}}0.48/\\ 0.43\end{tabular} &
  \begin{tabular}[c]{@{}c@{}}0.56/\\ 0.53\end{tabular} &
  \begin{tabular}[c]{@{}c@{}}0.60/\\ 0.68\end{tabular} &
  \begin{tabular}[c]{@{}c@{}}0.61/\\ 0.55\end{tabular} &
  54.41 &
  97.93 &
  105.32 &
  0.0706 &
  0.4558\\ \midrule
MAE\cite{he2022masked} &
  \begin{tabular}[c]{@{}c@{}}0.58/\\ 0.62\end{tabular} &
  \begin{tabular}[c]{@{}c@{}}0.54/\\ 0.48\end{tabular} &
  \begin{tabular}[c]{@{}c@{}}0.68/\\ 0.76\end{tabular} &
  \begin{tabular}[c]{@{}c@{}}0.53/\\ 0.61\end{tabular} &
  \begin{tabular}[c]{@{}c@{}}0.52/\\ 0.44\end{tabular} &
  \begin{tabular}[c]{@{}c@{}}0.61/\\ 0.39\end{tabular} &
  \begin{tabular}[c]{@{}c@{}}0.60/\\ 0.73\end{tabular} &
  \begin{tabular}[c]{@{}c@{}}0.48/\\ 0.79\end{tabular} &
  \begin{tabular}[c]{@{}c@{}}0.38/\\ 0.06\end{tabular} &
  65.10 &
  303.49 &
  329.23 &
  0.0614 &
  0.5928\\ \midrule

SwinMAE\cite{ma2023efficient} &
  \begin{tabular}[c]{@{}c@{}}0.64/\\ 0.63\end{tabular} &
  \begin{tabular}[c]{@{}c@{}}0.65/ \\ 0.68\end{tabular} &
  \begin{tabular}[c]{@{}c@{}}0.84/ \\ 0.83\end{tabular} &
  \begin{tabular}[c]{@{}c@{}}0.72/ \\ 0.61\end{tabular} &
  \begin{tabular}[c]{@{}c@{}}0.66/ \\ 0.70\end{tabular} &
  \begin{tabular}[c]{@{}c@{}}0.64/ \\ 0.48\end{tabular} &
  \begin{tabular}[c]{@{}c@{}}0.60/ \\ 0.74\end{tabular} &
  \begin{tabular}[c]{@{}c@{}}0.65/\\ 0.79\end{tabular} &
  \begin{tabular}[c]{@{}c@{}}0.68/\\ 0.62\end{tabular} &
  67.24 &
  26.93 &
  43.37 &
  0.0699 &
  0.6135 \\ \midrule
HIPT\cite{chen2022scaling} &
  \begin{tabular}[c]{@{}c@{}}0.72/\\ 0.78\end{tabular} &
  \begin{tabular}[c]{@{}c@{}}0.77/\\ 0.73\end{tabular} &
  \begin{tabular}[c]{@{}c@{}}0.79/\\ 0.83\end{tabular} &
  \begin{tabular}[c]{@{}c@{}}0.84/\\ 0.81\end{tabular} &
  \begin{tabular}[c]{@{}c@{}}0.84/\\ 0.82\end{tabular} &
  \begin{tabular}[c]{@{}c@{}}\textbf{0.78}/\\ 0.50\end{tabular} &
  \begin{tabular}[c]{@{}c@{}}0.63/\\ \textbf{0.84}\end{tabular} &
  \begin{tabular}[c]{@{}c@{}}0.56/\\ 0.79\end{tabular} &
  \begin{tabular}[c]{@{}c@{}}0.57/\\ 0.38\end{tabular} &
  67.46 &
  27.66 &
  - &
  0.0569&
  0.6724\\ \midrule
%% KAT zheng2023kernel SwinMAE ma2023efficient UVCGAN torbunov2023uvcgan 

KAT\cite{zheng2023kernel} &
  \begin{tabular}[c]{@{}c@{}}0.62/\\ 0.69\end{tabular} &
  \begin{tabular}[c]{@{}c@{}}0.75/ \\ 0.68\end{tabular} &
  \begin{tabular}[c]{@{}c@{}}0.80/ \\ 0.81\end{tabular} &
  \begin{tabular}[c]{@{}c@{}}0.87/ \\ 0.82\end{tabular} &
  \begin{tabular}[c]{@{}c@{}}0.82/ \\ 0.87\end{tabular} &
  \begin{tabular}[c]{@{}c@{}}0.68/ \\ 0.57\end{tabular} &
  \begin{tabular}[c]{@{}c@{}}0.63/ \\ 0.71\end{tabular} &
  \begin{tabular}[c]{@{}c@{}}0.63/\\ 0.70\end{tabular} &
  \begin{tabular}[c]{@{}c@{}}0.66/\\ 0.61\end{tabular} &
  71.79 &
  62.19 &
  80.57 &
  0.0721 &
  0.6837\\ \midrule

UVCGAN\cite{torbunov2023uvcgan} &
  \begin{tabular}[c]{@{}c@{}}0.66/\\ 0.74\end{tabular} &
  \begin{tabular}[c]{@{}c@{}}0.76/ \\ 0.69\end{tabular} &
  \begin{tabular}[c]{@{}c@{}}0.76/ \\ 0.82\end{tabular} &
  \begin{tabular}[c]{@{}c@{}}0.85/ \\ 0.77\end{tabular} &
  \begin{tabular}[c]{@{}c@{}}0.81/ \\ 0.84\end{tabular} &
  \begin{tabular}[c]{@{}c@{}}0.66/ \\ 0.54\end{tabular} &
  \begin{tabular}[c]{@{}c@{}}0.64/ \\ 0.70\end{tabular} &
  \begin{tabular}[c]{@{}c@{}}0.67/\\ 0.77\end{tabular} &
  \begin{tabular}[c]{@{}c@{}}0.70/\\ 0.62\end{tabular} &
  72.34 &
  32.19 &
  90.57 &
  0.0689 &
  0.6886\\ \midrule
DiffAE\cite{preechakul2022diffusion} &
  \begin{tabular}[c]{@{}c@{}}\textbf{0.73}/\\ \textbf{0.79}\end{tabular} &
  \begin{tabular}[c]{@{}c@{}}0.78/\\ \textbf{0.74}\end{tabular} &
  \begin{tabular}[c]{@{}c@{}}0.83/\\ 0.88\end{tabular} &
  \begin{tabular}[c]{@{}c@{}}0.92/\\ 0.88\end{tabular} &
  \begin{tabular}[c]{@{}c@{}}\textbf{0.87}/\\ 0.87\end{tabular} &
  \begin{tabular}[c]{@{}c@{}}0.69/\\ \textbf{0.55}\end{tabular} &
  \begin{tabular}[c]{@{}c@{}}0.63/\\ 0.74\end{tabular} &
  \begin{tabular}[c]{@{}c@{}}0.63/\\ 0.74\end{tabular} &
  \begin{tabular}[c]{@{}c@{}}0.68/\\ 0.58\end{tabular} &
  75.48 &
  25.59 &
  131.95 &
  0.0697 &
  0.7624\\ \midrule
PosDiffAE &
  \begin{tabular}[c]{@{}c@{}}0.72/\\ 0.77\end{tabular} &
  \begin{tabular}[c]{@{}c@{}}\textbf{0.80}/\\ 0.73\end{tabular} &
  \begin{tabular}[c]{@{}c@{}}\textbf{0.84}/\\ \textbf{0.89}\end{tabular} &
  \begin{tabular}[c]{@{}c@{}}\textbf{0.94}/\\ \textbf{0.89}\end{tabular} &
  \begin{tabular}[c]{@{}c@{}}\textbf{0.87}/\\ \textbf{0.90}\end{tabular} &
  \begin{tabular}[c]{@{}c@{}}0.75/\\ 0.53\end{tabular} &
  \begin{tabular}[c]{@{}c@{}}\textbf{0.65}/\\ 0.81\end{tabular} &
  \begin{tabular}[c]{@{}c@{}}\textbf{0.68}/\\ \textbf{0.80}\end{tabular} &
  \begin{tabular}[c]{@{}c@{}}\textbf{0.72}/\\ \textbf{0.63}\end{tabular} &
  \textbf{78.01} &
  25.59 &
  131.97 &
  0.0691 &
  0.7736 \\ \bottomrule
\end{tabular}
}
\label{table:Main_classif_table}
\end{table*}

%%%%%%%%%%%%%%%%%%%%%%%%%%%%%%%%%%%%%%%%%%%%%

\section{Experimental Setup}
\label{sec:exp}
\subsection{Dataset}
 
The datasets utilized here are obtained from (i) Allen BrainSpan\footnote{\href{http://brainspan.org}{http://brainspan.org}} and (ii) our in-house data repository. These datasets consist of HR Nissl-stained histological brain sections from the developmental stages of human brains. We have selected a group of regions for discrimination, ensuring good coverage across the brain. 
% A certain number of regions is chosen for classification to ensure that there is maximal coverage across brain regions visible in the coronal plane.

\textit{Allen Data:} From the Allen BrainSpan repository, we have utilized Nissl data of two fetal brains and one adult human brain corresponding to mid-gestational ages $21$, $15$ post-conceptual weeks (pcw) and $34$ years old, respectively. The digitization of sections was done at a resolution of $1\,\mu m$ per pixel with a thickness of $20$ $\mu m$ and $50$ $\mu m$ per section for fetal and adult brains, respectively. The plane of sectioning of the brain was coronal. The annotations in the $21$ pcw were done at intervals of $0.5$ - $1.2$$\, mm $ in $81$ sections of the cerebrum of the right hemisphere of the brain. Similarly, annotations were done for $15$ pcw and $34$-year-old adult brains at intervals of $0.5$ - $1.0$$\, mm $ in $46$ sections and $0.4$ - $3.4$$\, mm $ in $106$ sections, respectively. The selected regions cover $88.15\%$ of the $21$ pcw brain as indicated in Figure \ref{fig:Regions_present}. The regions for fetal brains are (i) cortical plate ({CP/ 10515}), (ii) subplate zone ({SP/ 10522}), (iii) sub ventricular zone ({SZ/ 10536}) (iv) ventricular zone ({VZ/ 10542}), (v) caudal ganglionic eminence ({CGE/ 10552}), (vi) caudate nucleus ({Ca/ 10334}), (vii) putamen ({Pu/ 10338}), (viii) corpus callosum ({cc/ 10561}), (ix) internal capsule ({ic/ 10581}).
The sections considered for fetal brains include transient structures that may disappear after the development process of the brain is finished, so the regions considered for the adult brain are (i) frontal neocortex, (ii) parietal neocortex, (iii) temporal neocortex, (iv) lateral nuclear complex of thalamus, (v) caudate nucleus, (vi) putamen and two layers of the hippocampus 
(vii) the stratum lacunosum-moleculare, and (viii) the stratum pyramidale.

\textit{In-house Data:} We have used the histological data of two de-identified fetal brains of $19$ and $23$ pcw obtained from the Department of Pathology, Mediscan Systems Pvt. Ltd Chennai, India. 
The sample processing and imaging of human brain tissue were conducted in accordance with the guidelines approved by the Institutional Ethics Committee (IEC) of the Indian Institute of Technology Madras (IITM) (Ethics Approval No. IEC/2021–01/MS/06).
The sections were acquired at a resolution of $0.5\,\mu m$ per pixel with a thickness of 20 $\mu m$ per section and cut in the coronal and sagittal plane for $19$ and $23$ pcw, respectively. We have selected nine regions mentioned in the Allen set, and additionally included the intermediate zone ({IZ / 10529}) region and excluded VZ.

%%%%%%%%%%%%%%%%%%%%%%%%%%%%%%%%%%%%%%%%%%%%%

\subsection{Evaluation Metrics}
\label{subsec:Evaluation Metrics}
\textit{Classification}: To evaluate the classification performance of the $classifier$ built on $PosDiffAE$ for different brain regions, we consider Precision, Recall, and Accuracy. We perform statistical analysis by evaluating Cohen's Kappa ($\kappa$).
\textit{Regression}: To compare the regressive capability of $PosDiffAE$ with other models which implicitly do not have this capability, we have trained Linear Regression models\footnote{\href{https://scikit-learn.org/stable/modules/generated/sklearn.linear_model.LinearRegression.html}{scikit-learn Linear Regression}} ($Model Name+Rg$) on the latent vectors of different models. Additionally, to quantify and statistically analyze the performance in the regression task, we have used Mean Squared Error (MSE) and a paired t-test between our method and the best-performing baseline. 
\textit{Restoration}: To verify whether the tear artifact restored images have preserved cellular properties like densities and count, we extracted these properties using a cell detection algorithm called Cellpose \cite{stringer2021cellpose}. 
Using Cellpose, we assessed cellular characteristics through cell occupancy (proportion of area occupied by detected cells) and cell count (number of detected cells) within image patches. We then calculated the relative absolute error of these metrics between the original and restored datasets.
To evaluate the distributional similarity, the commonly used Fr\'echet Inception Distance (FID) will not be suitable for histology data because the cellular image data may not align with the training data distribution of the Inception model. Thus, we estimate the Fr\'echet Distance between the latent vectors of the Cellpose model (FCD). In the context of JPEG restoration, we include peak signal-to-noise ratio (PSNR), structural similarity index metric (SSIM), and FCD. 
For statistical analysis, we performed a Wilcoxon rank-sum test between our method and the best-performing baseline. Statistically significant (from Wilcoxon and t-test) differences between our method and the best-performing baseline are marked with an asterisk ($^*$).

\subsection{Implementation Details and Comparison Methods}
\subsubsection{Data Details} \textit{Train Set} - We have used $7200$ image patches from $41$ out of $81$ brain sections in the $21$ pcw Allen brain. These 41 sections constituted every alternate section, starting from the first section. \textit{Validation Set} - The remaining $40$ sections were used for validation. This split ensures that both sets have representative patches from the regions of interest, reducing potential bias. In both cases, patches were considered if they completely belonged to any of the selected nine regions.
\textit{Cross-Data Setups/ Test Sets}- For the generalization experiment, the $PosDiffAE$ being trained on $21$ pcw, we have tested the performance on image patches from $15$ pcw, $19$ pcw, $23$ pcw, and $34$-year-old adult brain. 
The pre-processing step applied to all the patches was a linear intensity normalization within the range of [0, 1]. The in-house $19$ pcw brain images were acquired at $0.5\,\mu m$ per pixel, so we have taken image patches of $512 \times 512$ and down-sampled to $256 \times 256$. We performed normalization in the range of [0, 1] on the radial distance and angle. The \textit{Original Set} referred to in the experiments is generated with $10000$ patches from the $21$ pcw Allen validation set. 
In the $21$ pcw validation set brain images, a few regions appeared blank but were marked as valid brain regions in the ground truth annotations.
Blank patches were extracted from these regions, forming the \textit{Tear Artifact} set with a count of $5000$ patches. These patches were excluded from the \textit{Original Set}. The \textit{JPEG Compression} set was created on the $21$ pcw Allen validation set by applying three levels of degradation corresponding to Quality Factor (QF) $5$, $15$, and $25$, respectively. Similarly, the \textit{Gamma Adjustment} set was created on $21$ pcw Allen validation set by varying gamma adjustment values $\gamma$ = $1.2$, $1.3$, and $1.4$. Similar to the three sets for the $21$ pcw brain, these three sets were created for the Cross-Data Setups/test datasets ($23$ pcw and $34$-year-old brains). Additionally, a \textit{Black-dot} artifact set was simulated for $23$ pcw brain patches (test set). These black dots, caused by unwanted substances during image acquisition, were simulated using the cut-and-paste method, as they fully occlude tissue regions.

\subsubsection{Model Details} The models were implemented in PyTorch version 2.0.1 on an 80 GB NVIDIA A100 GPU and CUDA Version: 12.1. The additional parameter details are listed in Table \ref{table:parameter_details}. The baselines chosen for classification and regression are the following set of recent transformer-based models: Trans-UNet \cite{chen2021transunet}, MAE \cite{he2022masked}, SwinMAE \cite{ma2023efficient}, HIPT \cite{chen2022scaling}, KAT \cite{zheng2023kernel}. Additionally, generative modeling approaches, a GAN-based transformer model (UVCGAN \cite{torbunov2023uvcgan}) and a diffusion auto-encoding model (DiffAE \cite{preechakul2022diffusion}) are included as baselines. An image inpainting-based task was used to train the UVCGAN model, keeping the architectural specification and adversarial loss component similar to the original implementation. The KAT model is implemented on the features of the SwinMAE since these models are better feature extractors as compared to CNNs in the original implementation.
For restoration with the \textit{Tear Artifact} set, we have considered an ablated version of our model ($PosDiffAE+Restore\_wo$) without $Enc_{\phi^{'}}$ and therefore resembling the conventional DDM-based in-painting models \cite{lugmayr2022repaint}. We have also included diffusion-based models ArtiFusion \cite{he2023artifact} and HARP \cite{fuchs2024harp} as baselines for restoration.
For JPEG restoration along with $PosDiffAE+Restore\_wo$, we have compared with a blind CNN technique, FBCNN \cite{jiang2021towards}, and a non-blind DDM-based approach, DDRM \cite{kawar2022denoising}. These models were implemented following the official code repositories and adapting to our requirements. We make our evaluation codes along with additional information, available at \href{https:/github.com/ayantikadas/PosDiffAE_}{https:/github.com/ayantikadas/PosDiffAE\_}.

%%%%%%%%%%%%%%%%%%%%%%%%%%%%%%%%%%%%%%%
  %   \cite{chen2022scaling} 
  % \cite{chen2021transunet}, \cite{he2022masked}, \cite{preechakul2022diffusion} 
%% KAT zheng2023kernel SwinMAE ma2023efficient UVCGAN torbunov2023uvcgan 

\begin{table}[!hbt]
\caption{Analysis of the classification performance of different models under multiple artifact settings ($21$ pcw validation dataset).}
\centering
\resizebox{\columnwidth}{!}{%
% \resizebox{0.8\textwidth}{!}{%
\begin{tabular}{@{}lllllllll@{}}
\toprule
\textbf{Methods} & \multicolumn{8}{c}{\textbf{Accuracy ($\uparrow$)(\%)}}                    \\ \midrule
                  & \multicolumn{1}{c}{\begin{tabular}[c]{@{}c@{}}Original \\ Set\end{tabular}}    & \multicolumn{1}{c}{\begin{tabular}[c]{@{}c@{}}Tear \\ Artifact\end{tabular}}  & \multicolumn{3}{l}{JPEG Compression }   & \multicolumn{3}{l}{Gamma Adjustment}             \\
                 &       &       & QF 5 & QF 10 & QF 15 & 1.2   & 1.3   & 1.4   \\ \midrule
Trans-UNet\cite{chen2021transunet}       & 54.41 & 52.44 & 44.06 & 53.01 & 53.01 & 44.76 & 39.87 & 36.63 \\ \midrule
MAE\cite{he2022masked}              & 65.10 & 62.60 & 40.88 & 58.79 & 62.03 & 61.07 & 58.34 & 55.42 \\ \midrule

SwinMAE \cite{ma2023efficient}             & 67.24 & 65.80 & 30.26 & 39.40 & 43.20 & 61.73 & 58.33 & 53.33 \\ \midrule

HIPT\cite{chen2022scaling}             & 67.46 & 64.50 & 45.20 & 60.95 & 63.80 & 62.34 & 60.19 & 56.44 \\ \midrule

KAT\cite{zheng2023kernel}             & 71.79 & 67.39 & 34.19 & 43.53 & 46.86 & 65.99 & 61.53 & 56.33 \\ \midrule

UVCGAN\cite{torbunov2023uvcgan}        & 72.34 & 59.46 & 58.53 & 54.66 & 58.26 & 64.60 & 60.60 & 56.66 \\ \midrule

DiffAE\cite{preechakul2022diffusion}            & 75.48 & 59.74 & 66.66 & 72.43 & 73.65 & 70.22 & 65.46 & 60.88 \\ \midrule
PosDiffAE         & \textbf{78.01} & 63.42          & 68.19          & 72.50          & \textbf{74.47} & \textbf{72.88} & \textbf{69.50} & \textbf{66.34} \\ \midrule
\begin{tabular}[c]{@{}l@{}}PosDiffAE+\\ Restore\end{tabular} &
  -              & \textbf{71.17} & \textbf{68.63} & \textbf{72.56} & 74.07          & -              & -              & -              \\ \bottomrule
\end{tabular}%
}
% \end{table}
\label{table:Atifcat_classif_table}
\end{table}
%%%%%%%%%%%%%%%%%%%%%%%%%%%%%%%%%%%%

\section{Results and Discussion}
The results of experimentation with the validation set from $21$ pcw brain and the test sets from 15, 19, $23$ pcw fetal and $34$-year-old adult brain are demonstrated and inferred in the subsections below.
\subsection{Quantitative Evaluation}
\subsubsection{Classification of Brain Regions} The quantitative results for the classification of brain regions are indicated in Table \ref{table:Main_classif_table}, based on the quantifiers (precision/ Recall and Accuracy). The accuracy indicates that our proposed method $PosDiffAE$ outperforms other baselines in overall region average accuracy values. 
%%%%%%%%%%%%%%%%
% Increase in precision and Increase in Recall:
The addition of positional information ($r_{0}$ and $\theta_{0}$) has clearly aided in reducing the False Negative (FN) and False Positive (FP) rates in the regions of the SZ, SP, and CP. This reduction is attributed to the reduction in inter-class confusion between SP and SZ, as well as the distinct variation in inter-class positional information between CP and VZ, thereby decreasing the misclassification rate of VZ as CP. Similarly, the addition of positional information attributed to the increment of precision and recall scores for the cc and ic.
%%%%%%%%%%%%%%%%
% Increase in precision and decrease in Recall:
For the Pu and CGE regions, the precision score has increased due to the reduction in FP rates. However, the performance of $DiffAE$ surpasses that of $PosDiffAE$ in the case of Ca, indicating that the incorporation of positional information does not enhance the performance. This is primarily because both Pu and Ca regions share spatial adjacency, leading to similar positional information in terms of $r_{0}$ and $\theta_{0}$. 
From the classification results, it is inferred that regions which have distinct cellular properties (CP, SP, SZ) are better separable as compared to classes which exhibit similar cellular density and morphological properties (cc and ic, VZ and CGE, Ca and Pu) at a patch level. 

% can be better distinguished by $PosDiffAE$ due to positional awareness.
%%%%%%%%%%%%%%%%
There is an incremental nature of evaluation metrics for DDM models in the case of sparse and mid-dense cellular regions (as in Table \ref{table:Main_classif_table}). However, this nature of performance does not extend to highly dense regions like the Ventricular Zone (VZ). In such high-density areas, the transformer-based HIPT \cite{chen2022scaling} model outperforms DDM-based models due to its superior ability to capture global contextual correlations between the pixels of a patch. 
While the UVCGAN model does not explicitly exhibit superiority in dense classes, it shows improvement in medium-density and sparse classes, owing to an overall better accuracy among all the transformer-based baselines. The KAT model performs better than all other transformer-based models except UVCGAN, owing to its awareness of the neighborhood patches. The SwinMAE model performs better than the MAE model since the windowing approach of Swin-transformers can capture better contextual information within an image patch. 
The MAE \cite{he2022masked}  and Trans-UNet \cite{chen2021transunet} models, while also based on the transformer backbones, show relatively lower metric values. The number of parameters and the inference time requirement for each of the models are indicated in  Table \ref{table:Main_classif_table}. Among the encoding parameters required during the classification, the diffusion models have the least number of parameters involved. For the AE training, the SwinMAE model has the lowest number of parameters.  

%%%%%%%%%%%%%%%%%%%%%%%%%%%%%%%%%%%%%%
\begin{figure}
  \centering
  \centerline{\includegraphics[width=1\linewidth]{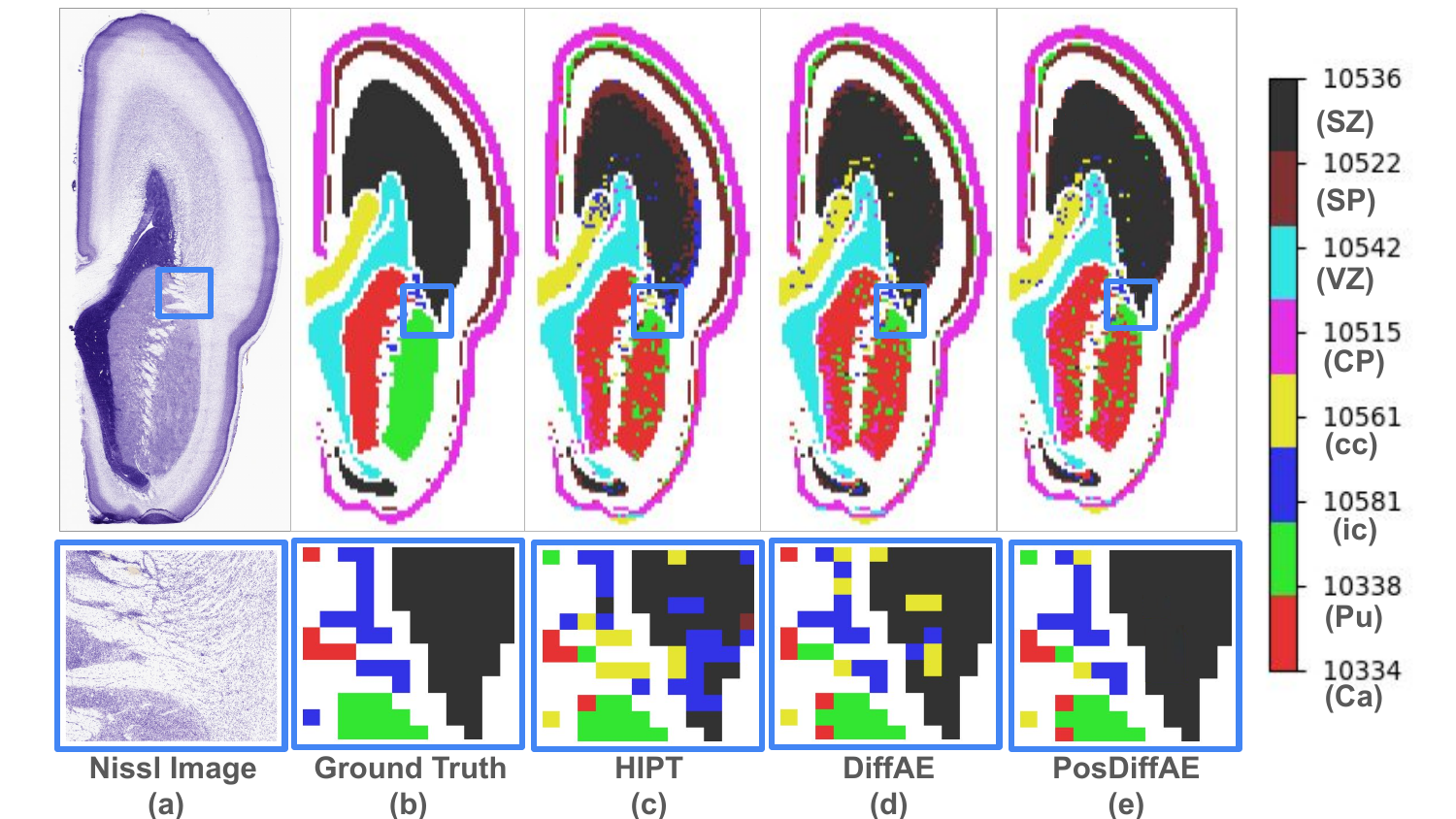}}
\caption{Region classification for a section from $21$ pcw Allen brain (validation set). From left to right, (a) Nissl image section, (b) Ground Truth annotations, Predictions form (c) HIPT, (d) DiffAE, (e) PosDiffAE. The top row indicates the whole section of the brain, and the bottom row highlights a specific region from the section. The blue box shows that PosDiffAE has relatively fewer misclassifications in a region surrounded by four different classes.}
\label{fig:Classification}
\end{figure}
%%%%%%%%%%%%%%%%%%%%%%%%%%%%%%%%%%%%%%
%%%%%%%%%%%%%%%%%%%%%%%%%%%%%%%%%%%%%%%
\begin{figure}[]
  \centering
  \centerline{\includegraphics[width=1.0\linewidth]{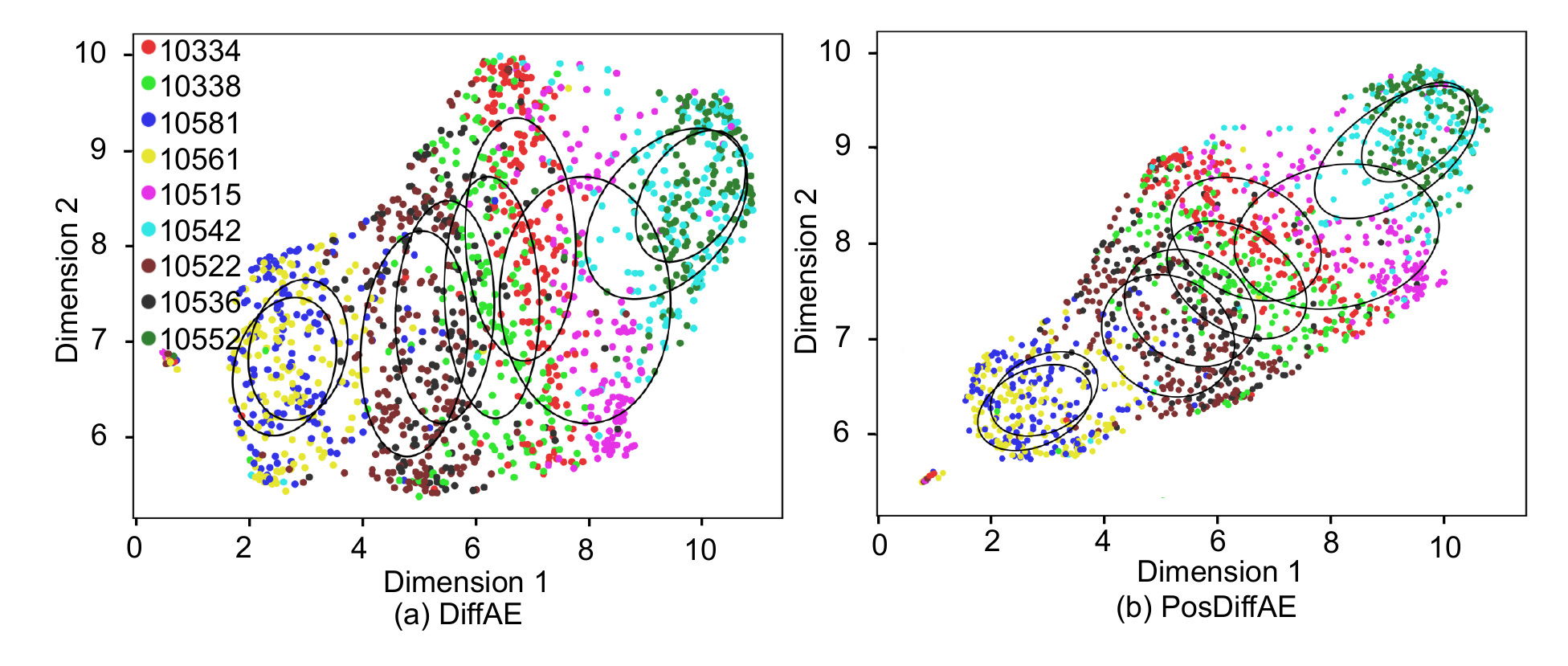}}
\caption{Umap-based $2\mathrm{D}$ projections of the latent representations of (a) DiffAE and (b) PosDiffAE for $21$ pcw (validation set). We extracted the ellipses from the eigen-components of the covariance matrix of regions.}
\label{fig:Latent_space}
\end{figure}
%%%%%%%%%%%%%%%%%%%%%%%%%%%%%%%%%%%%%

%%%%%%%%

\subsubsection{Classification Under Artifact Settings}
The classification accuracy of different methods under multiple artifact settings is quantified in Table \ref{table:Atifcat_classif_table}. The $PosDiffAE$ model exhibits better accuracy compared to the DDM baseline model ($DiffAE$) and other models based on the transformer backbone. When the classification is evaluated on \textit{Tear Artifact}, characterized by unwanted white regions within the image patches, there is an overall decrease in accuracy for all methods. However, our restoration method ($PosDiffAE+Restore$) shows improvements in performance, suggesting that the restored images have incorporated contextually meaningful content within the tear regions in a completely unsupervised manner. For these tear artifact images, the transformer models (SwinMAE, KAT, and HIPT) demonstrate better accuracy. Specifically, the KAT and SwinMAE models, trained using masking strategies resembling tear artifacts, can derive meaningful features out of missing context. The performance drop in DDM models is attributed to the robust and semantically meaningful representation capabilities of these models, which are affected due to the presence of tear artifacts in the images, leading to relatively lower accuracy. Similarly, for \textit{JPEG Compression} set, there is an overall reduction in performance for all methods. The performance in terms of accuracy correlates negatively with the level of compression. However, within varying levels of degradation, the DDM set of models performs better than others, with $PosDiffAE$ demonstrating superior performance across all cases. Despite undergoing distortion due to compression, the $PosDiffAE$ model maintains better performance as there is no alteration in the semantic content of the images. For higher levels of compression (QF 5 and 10) $PosDiffAE+Restore$ technique could yield better classification accuracy. However, for SwinMAE and KAT, JPEG compression notably impacts performance as these models depend heavily on high-frequency details in the unmasked regions for feature extraction during training. Additionally, for \textit{Gamma Adjustment}, set $PosDiffAE$ has higher rates of accuracy, highlighting its resilience to perturbations and changes that do not affect the semantic content of the images.

\subsection{Qualitative Evaluation}
The qualitative results for patch-wise classification in a brain section from the $21$ pcw Allen data are depicted in Figure \ref{fig:Classification}, showcasing eight observable classes. Within the $PosDiffAE$ model, the SZ region (black) has fewer misclassifications with the SP region (brown) compared to other baselines. For $PosDiffAE$ in the regions between CP (pink) and SP (brown), the number of misclassified SP patches as Pu (green) is relatively low.
% The blue bounding box in Figure \ref{fig:Classification} emphasizes that in multi-class regions, $PosDiffAE$ performs better than other baselines, particularly with fewer SZ misclassifications. The HIPT model shows more SZ misclassifications compared to diffusion-based models.
The blue bounding box in Figure \ref{fig:Classification} highlights that in multi-class regions, $PosDiffAE$ outperforms other baselines, particularly by reducing SZ misclassifications. The HIPT model, in contrast, shows more SZ misclassifications compared to diffusion-based models. As indicated in the blue box, the HIPT model tends to misclassify SZ regions as ic (blue) in certain regions. Additionally, in the upper SZ regions, HIPT has higher misclassification rates of SZ as SP (brown). Notably, misclassifications arise between visually similar regions such as ic (dark blue) and cc (yellow), and Pu (green) and Ca (red), respectively. Furthermore, transitions between regions exhibit misclassifications primarily due to gradient changes rather than steep transitions. Defects during acquisition, such as folded regions at the bottom of the Nissl image in Figure \ref{fig:Classification} within region CP, confuse the classification model despite the ground truth annotation marking it as a complete CP region.

% Trans-UNet\cite{chen2021transunet} 

% MAE\cite{he2022masked}

% HIPT\cite{chen2022scaling}

% DiffAE\cite{preechakul2022diffusion}
% \usepackage{booktabs}
% \usepackage{graphicx}
\begin{table}[]
\caption{Analysis of the regressive capabilities. We report the MSE metric between the original and predicted $r$/$\theta$, and $^*$ denotes statistical significance ($p<0.01$).}
\centering
\resizebox{\columnwidth}{!}{%
\begin{tabular}{@{}lllllllll@{}}
\toprule
\multicolumn{1}{c}{\textbf{Methods}} &
  \multicolumn{8}{c}{\textbf{MSE $r$ / MSE $\theta$} ( $10^{-2}$) $\downarrow$} \\ \midrule
\multicolumn{1}{c}{} &
  \multicolumn{1}{c}{\begin{tabular}[c]{@{}c@{}}Original \\ Set\end{tabular}} &
  \multicolumn{1}{c}{\begin{tabular}[c]{@{}c@{}}Tear \\ Artifact\end{tabular}} &
  \multicolumn{3}{c}{JPEG Compression} &
  \multicolumn{3}{c}{Gamma Adjustment} \\ \midrule
\multicolumn{1}{c}{} &
  \multicolumn{1}{c}{} &
  \multicolumn{1}{c}{} &
  \multicolumn{1}{c}{QF 5} &
  \multicolumn{1}{c}{QF 10} &
  \multicolumn{1}{c}{QF 15} &
  \multicolumn{1}{c}{1.2} &
  \multicolumn{1}{c}{1.3} &
  \multicolumn{1}{c}{1.4} \\ \midrule
TransUNet+Rg &
  \begin{tabular}[c]{@{}l@{}}7.28/ \\ 7.39\end{tabular} &
  \begin{tabular}[c]{@{}l@{}}7.36/\\ 7.50\end{tabular} &
  \begin{tabular}[c]{@{}l@{}}7.57/\\ 8.62\end{tabular} &
  \begin{tabular}[c]{@{}l@{}}7.37/ \\ 7.55\end{tabular} &
  \begin{tabular}[c]{@{}l@{}}7.31/ \\ 7.49\end{tabular} &
  \begin{tabular}[c]{@{}l@{}}7.94/\\ 7.39\end{tabular} &
  \begin{tabular}[c]{@{}l@{}}8.29/ \\ 7.49\end{tabular} &
  \begin{tabular}[c]{@{}l@{}}8.72/\\ 7.73\end{tabular} \\ \midrule
MAE+Rg &
  \begin{tabular}[c]{@{}l@{}}7.13/\\ 7.09\end{tabular} &
  \begin{tabular}[c]{@{}l@{}}8.31/\\ 7.63\end{tabular} &
  \begin{tabular}[c]{@{}l@{}}7.83/ \\ 10.58\end{tabular} &
  \begin{tabular}[c]{@{}l@{}}7.32/ \\ 7.85\end{tabular} &
  \begin{tabular}[c]{@{}l@{}}6.88/ \\ 7.38\end{tabular} &
  \begin{tabular}[c]{@{}l@{}}9.09/\\ 8.01\end{tabular} &
  \begin{tabular}[c]{@{}l@{}}11.05/\\ 8.77\end{tabular} &
  \begin{tabular}[c]{@{}l@{}}14.67/ \\ 9.88\end{tabular} \\ \midrule
SwinMAE+Rg &
  \begin{tabular}[c]{@{}l@{}}5.97/\\ 6.58\end{tabular} &
  \begin{tabular}[c]{@{}l@{}}6.98/\\ 7.41\end{tabular} &
  \begin{tabular}[c]{@{}l@{}}11.88/\\ 9.62\end{tabular} &
  \begin{tabular}[c]{@{}l@{}}10.23/\\ 8.13\end{tabular} &
  \begin{tabular}[c]{@{}l@{}}9.63/ \\ 7.62\end{tabular} &
  \begin{tabular}[c]{@{}l@{}}8.05/\\ 6.84\end{tabular} &
  \begin{tabular}[c]{@{}l@{}}9.35/\\ 7.94\end{tabular} &
  \begin{tabular}[c]{@{}l@{}}10.29/\\ 8.82\end{tabular} \\ \midrule
HIPT+Rg &
  \begin{tabular}[c]{@{}l@{}}7.36/\\ 7.42\end{tabular} &
  \begin{tabular}[c]{@{}l@{}}7.02/\\ 7.41\end{tabular} &
  \begin{tabular}[c]{@{}l@{}}7.04/\\ 10.19\end{tabular} &
  \begin{tabular}[c]{@{}l@{}}6.06/\\ 6.90\end{tabular} &
  \begin{tabular}[c]{@{}l@{}}6.12/ \\ 6.76\end{tabular} &
  \begin{tabular}[c]{@{}l@{}}8.43/\\ 7.27\end{tabular} &
  \begin{tabular}[c]{@{}l@{}}10.99/\\ 8.61\end{tabular} &
  \begin{tabular}[c]{@{}l@{}}15.13/\\ 11.10\end{tabular} \\ \midrule
KAT+Rg &
  \begin{tabular}[c]{@{}l@{}}5.68/\\ 6.16\end{tabular} &
  \begin{tabular}[c]{@{}l@{}}6.91/\\ 7.30\end{tabular} &
  \begin{tabular}[c]{@{}l@{}}10.70/\\ 9.56\end{tabular} &
  \begin{tabular}[c]{@{}l@{}}9.40/\\ 8.55\end{tabular} &
  \begin{tabular}[c]{@{}l@{}}8.32/ \\ 7.77\end{tabular} &
  \begin{tabular}[c]{@{}l@{}}6.33/\\ 6.90\end{tabular} &
  \begin{tabular}[c]{@{}l@{}}7.13/ \\ 7.64\end{tabular} &
  \begin{tabular}[c]{@{}l@{}}8.63/\\ 8.04\end{tabular} \\ \midrule
UVCGAN+Rg &
  \begin{tabular}[c]{@{}l@{}}5.37/\\ 6.24\end{tabular} &
  \begin{tabular}[c]{@{}l@{}}6.81/\\ 7.16\end{tabular} &
  \begin{tabular}[c]{@{}l@{}}7.22/\\ 8.79\end{tabular} &
  \begin{tabular}[c]{@{}l@{}}7.01/\\ 7.04\end{tabular} &
  \begin{tabular}[c]{@{}l@{}}6.60/ \\ 6.67\end{tabular} &
  \begin{tabular}[c]{@{}l@{}}7.11/\\ 6.79\end{tabular} &
  \begin{tabular}[c]{@{}l@{}}7.40/\\ 7.86\end{tabular} &
  \begin{tabular}[c]{@{}l@{}}7.59/\\ 7.98\end{tabular} \\ \midrule
DiffAE+Rg &
  \begin{tabular}[c]{@{}l@{}}5.39/ \\ 5.99\end{tabular} &
  \begin{tabular}[c]{@{}l@{}}8.86/\\ 8.55\end{tabular} &
  \begin{tabular}[c]{@{}l@{}}6.72/\\ 8.63\end{tabular} &
  \begin{tabular}[c]{@{}l@{}}6.05/\\ 6.67\end{tabular} &
  \begin{tabular}[c]{@{}l@{}}6.00/\\ 6.66\end{tabular} &
  \begin{tabular}[c]{@{}l@{}}6.63/\\ 6.79\end{tabular} &
  \begin{tabular}[c]{@{}l@{}}8.01/\\ 7.29\end{tabular} &
  \begin{tabular}[c]{@{}l@{}}7.29/\\ 7.93\end{tabular} \\ \midrule
PosDiffAE &
  \begin{tabular}[c]{@{}l@{}}\textbf{5.24$^*$}/\\ \textbf{5.98$^*$}\end{tabular} &
  \begin{tabular}[c]{@{}l@{}}8.16$^*$/\\ 8.01$^*$\end{tabular} &
  \begin{tabular}[c]{@{}l@{}}5.50$^*$/\\ 8.62$^*$\end{tabular} &
  \begin{tabular}[c]{@{}l@{}}5.51$^*$/ \\ 6.63$^*$\end{tabular} &
  \begin{tabular}[c]{@{}l@{}}5.39$^*$/\\ 6.58$^*$\end{tabular} &
  \begin{tabular}[c]{@{}l@{}}\textbf{6.48$^*$}/\\ \textbf{6.75$^*$}\end{tabular} &
  \begin{tabular}[c]{@{}l@{}}\textbf{7.39$^*$}/\\ \textbf{7.21$^*$}\end{tabular} &
  \begin{tabular}[c]{@{}l@{}}\textbf{7.22$^*$}/\\ \textbf{7.69$^*$}\end{tabular} \\ \midrule
\begin{tabular}[c]{@{}l@{}}PosDiffAE+\\ Restore\end{tabular} &
  - &
  \begin{tabular}[c]{@{}l@{}}\textbf{6.76$^*$}/ \\ \textbf{7.02$^*$}\end{tabular} &
  \begin{tabular}[c]{@{}l@{}}\textbf{5.48$^*$}/\\ \textbf{8.50$^*$}\end{tabular} &
  \begin{tabular}[c]{@{}l@{}}\textbf{5.50$^*$}/ \\ \textbf{6.53$^*$}\end{tabular} &
  \begin{tabular}[c]{@{}l@{}}\textbf{5.27$^*$}/\\ \textbf{6.55$^*$}\end{tabular} &
  - &
  - &
  -

  \\ \bottomrule
\end{tabular}%
}
% \end{table}

\label{table:Artifact_r_theta}
\end{table}
%%%%%%%%%%%%%%%%%%%%%%%%%%%%%%%%

\subsection{Latent Space Analysis}

The comparison of the two-dimensional Umap \cite{mcinnes2018umap} projection of the latent vectors between the $DiffAE$ and proposed $PosDiffAE$ model is illustrated in Figure \ref{fig:Latent_space}. 
%
% These 2D projections, estimated using Umap, effectively capture both global and local relationships between the vectors. 
%
The Umap plots highlight that the arrangement of vectors in these $2\mathrm{D}$ projected spaces is dependent on the cellular-level properties and appearances. Vectors representing images with lower cellular density tend to cluster in regions characterized by lower values in both dimensions, while vectors representing images with higher cellular density exhibit increments in both dimensions. 
In Figure \ref{fig:Latent_space}, the variance of each of the regions along the two dimensions is represented by generating ellipses from the eigen-components of the covariance matrix of each region. 
% blue dots represent the center of the original 9 data clusters, while red dots represent the maximum meaningful cluster that could be formed in these 2D projections using k-means clustering.
%
From the plots, it is evident that the sparse regions cc (10561) and ic (10581), and the highly dense regions VZ (10542) and CGE (10552) have maximally overlapping ellipses. In the medium-density regions, the ellipses of (i) SZ (10536) and SP (10522) and (ii) Ca (10334) and Pu (10338) also show overlapping areas. Additionally, the spread of the ellipses indicates the frequency of occurrence and variability within certain regions. For instance, the CP (10515) region appears in multiple brain sections, resulting in a larger spread and greater variability in the CP ellipse.
%
% SZ (10536) and SP (10522), (iii) Ca (10334) and Pu (10338), (iv) CP 10515 . The movement from (i) to (v) represents a transition from sparse to dense groups of brain regions. 
The comparison of Figure \ref{fig:Latent_space} (a) and (b) suggests that the clusters in $PosDiffAE$ are more isotropic along the two directions of variations, indicating a more compact distribution of data points around the cluster centroids. This enhanced compactness aids in better separability of the brain regions in $PosDiffAE$.

%%%%%%%%%%%%%%%%%%%%%%%%%%% Image for  FB62
%%%%%%%%%%%%%%%%%%%%%%%%%%%%%%%%%%%%%%
\begin{figure}[!t]
  \centering
  \centerline{\includegraphics[width=1\linewidth]{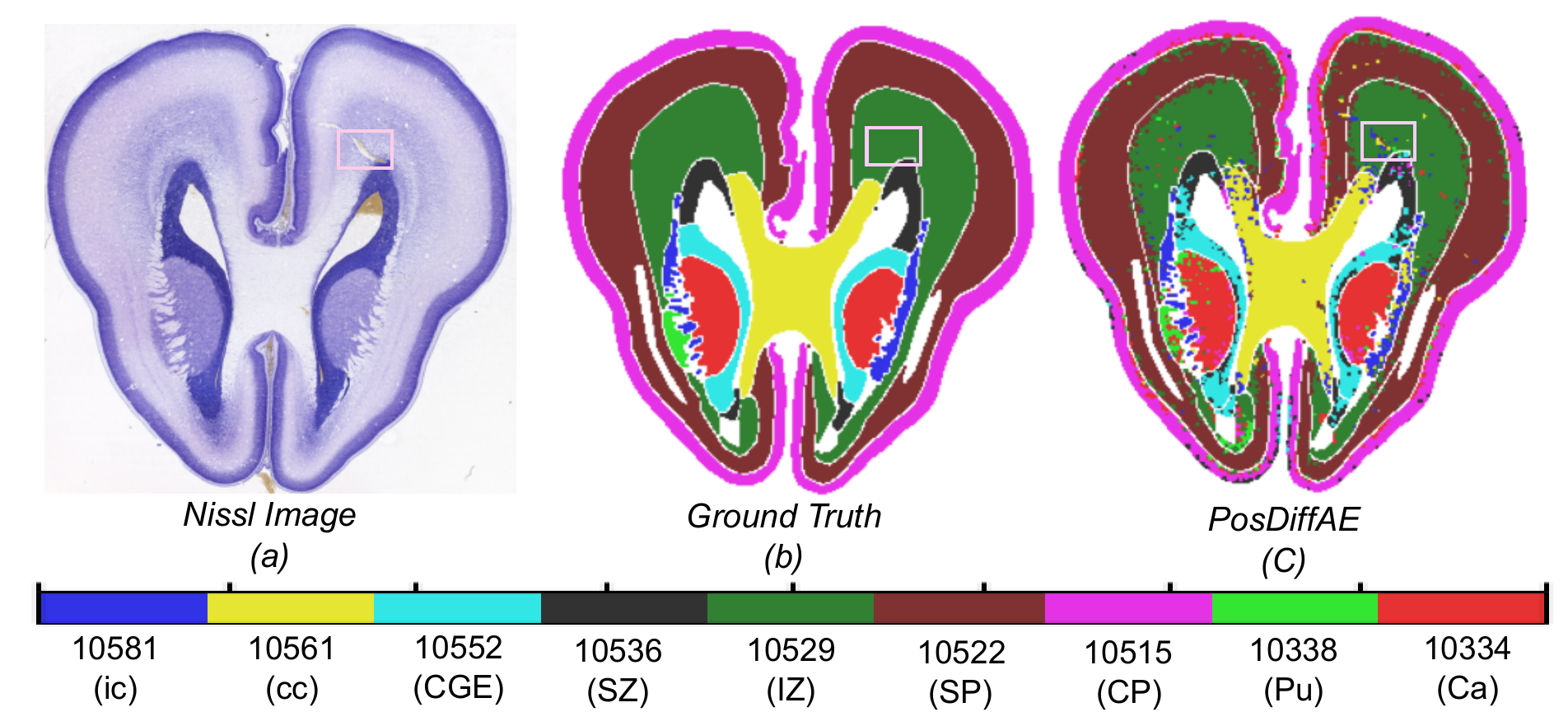}}
\caption{Classification of nine regions in a brain section from $19$ pcw. From left to right, (a) Nissl Image, (b) Ground Truth, and prediction are from (c) PosDIffAE. The pink box highlights that in an artifact region, the model is confused since the region's appearance deviates from IZ. }
\label{fig:FB62}
\end{figure}
%%%%%%%%%%%%%%%%%%%%%%%%%%%%%%%%%%%%%

% The latent space analysis was performed using Umap \cite{mcinnes2018umap}.

% %%%%%%%%%%%%%%%%%%%%%%%%%%%%%%%%%%%%%
% MSE $r$ / MSE $\theta$} ($10^{-2}$)
%%%%%%%%%%%%%%%%%%%%%%%%%%%%%%%%%%%%%%
\begin{figure*}[t]
  \centering
  \centerline{\includegraphics[width=1\linewidth]{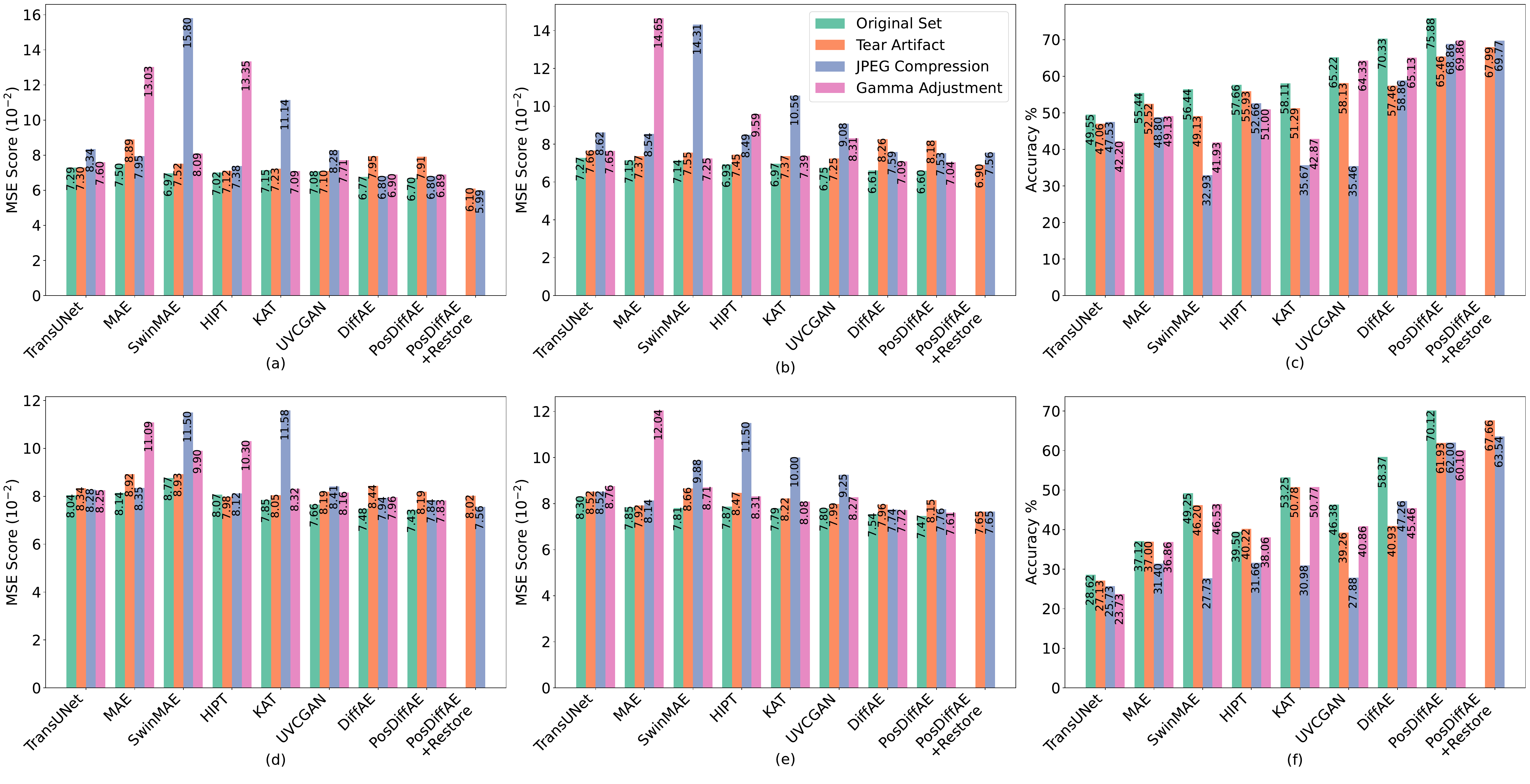}}
\caption{The top row, from left to right, represents the MSE errors of (a) $r$, (b) $\theta$, and the (c) classification accuracies for the original and various artifact data settings in $23$ pcw fetal brain (test dataset). Similarly, the bottom row denotes MSE errors (c), (d), and classification accuracies (e) for the $34$-year-old adult brain (test dataset). The original and various artifact sets are indicated by the four color coding. A lower height in the bars of the MSE plot indicates better performance, and for accuracy, a higher height denotes better performance.   }
\label{fig:cross_data_artif}
\end{figure*}
%%%%%%%%%%%%%%%%%%%%%%%%%%%%%%%%%%%%%

\subsection{Analysis Of Regressive Capabilities}

The regressive capabilities of different models are calculated as the MSE  between the predicted and original positional information and are listed in Table \ref{table:Artifact_r_theta}.
For the $PosDiffAE$ the radial positional information ($r$) and angular positional information ($\theta$) is extracted from the Radial Regressor ($f_{r}$) and the Angular Regressor ($f_{\theta^{'}}$) respectively. For other models that do not implicitly have regressive elements, these regression tasks are achieved using a Linear Regressor (Regress) as mentioned in Subsection \ref{subsec:Evaluation Metrics}. The regression task is performed on the latent representations of the original images and artifact images of all the models.
From Table \ref{table:Artifact_r_theta}, it's evident that the regression components of $PosDiffAE$ yield the lowest error values for both the original data settings and the artifact settings. Additionally, the error for regression of radial distance and angle correlates with the size of the artifact; as the artifact increases, so does the error. Notably, in the case of \textit{Tear Artifact} set, the error for the $PosDifAE$ model is higher since the artifact impacts the semantic content of the images. However, $PosDiffAE$ outperforms other models for \textit{JPEG Compression} and \textit{Gamma Adjustment} sets, where the semantics remain intact. Moreover, the latent representations of the restored images (using $PosDiffAE+Restore$) tend to perform better at the regressive task, similar to the classification task.  The transformer-based models (SwinMAE, KAT, and UVCGAN), which are trained to restore contents in the spatially masked regions, perform better for tear restoration. However, these models do not yield meaningful features from JPEG-compressed images, which causes a relatively higher drop in performance.

%%%%%%%%%%%%%%%%%%%%%%%%%%%%%%%%%%%%%%%%%%%%%%%%%%
\begin{table}[!t]
\caption{Overall classification Performance across all regions in Cross-Data Setups/test datasets and in terms of Precision (Pr.), Recall (Rec.), and Accuracy (Acc.) \%.}
\centering
\resizebox{\columnwidth}{!}{%
\begin{tabular}{@{}ccccccccc@{}}
\toprule
Methods & \multicolumn{8}{c}{Trained on $21$ pcw Allen brain} \\ \midrule
 & \multicolumn{2}{c|}{\begin{tabular}[c]{@{}c@{}}Inferred on \\ $15$ pcw brain ($\uparrow$)\end{tabular}} & \multicolumn{2}{c|}{\begin{tabular}[c]{@{}c@{}}Inferred on \\ $19$ pcw brain ($\uparrow$)\end{tabular}} & \multicolumn{2}{c|}{\begin{tabular}[c]{@{}c@{}}Inferred on \\ $23$ pcw brain ($\uparrow$)\end{tabular}} & \multicolumn{2}{c}{\begin{tabular}[c]{@{}c@{}}Inferred on \\ 34 year brain ($\uparrow$)\end{tabular}} \\ \midrule
 & \begin{tabular}[c]{@{}c@{}}Pr./\\ Rec.\end{tabular} & \multicolumn{1}{c|}{Acc.} & \begin{tabular}[c]{@{}c@{}}Pr./\\ Rec. \end{tabular} & \multicolumn{1}{c|}{Acc.} & \begin{tabular}[c]{@{}c@{}}Pr./\\ Rec.\end{tabular} & \multicolumn{1}{c|}{Acc.} & \begin{tabular}[c]{@{}c@{}}Pr./\\ Rec.\end{tabular} & Acc.\\ \midrule
TransUNet\cite{chen2021transunet} & \begin{tabular}[c]{@{}c@{}}0.36/\\ 0.37\end{tabular} & \multicolumn{1}{c|}{37.22} & \begin{tabular}[c]{@{}c@{}}0.61/\\ 0.65\end{tabular} & \multicolumn{1}{c|}{61.31} & {\begin{tabular}[c]{@{}c@{}}0.49/\\ 0.49\end{tabular}} & \multicolumn{1}{c|}{{49.55}} & \begin{tabular}[c]{@{}c@{}}0.28/\\ 0.27\end{tabular} & 28.62 \\ \midrule
MAE\cite{he2022masked} & \begin{tabular}[c]{@{}c@{}}0.41/\\ 0.36\end{tabular} & \multicolumn{1}{c|}{36.72} & \begin{tabular}[c]{@{}c@{}}0.65/\\ 0.67\end{tabular} & \multicolumn{1}{c|}{69.40} & \begin{tabular}[c]{@{}c@{}}0.52/\\ 0.53\end{tabular} & \multicolumn{1}{c|}{52.44} & \begin{tabular}[c]{@{}c@{}}0.37/\\ 0.37\end{tabular} & 37.12 \\ \midrule
SwinMAE\cite{ma2023efficient} & \begin{tabular}[c]{@{}c@{}}0.35/\\ 0.36\end{tabular} & \multicolumn{1}{c|}{37.29} & \begin{tabular}[c]{@{}c@{}}0.71/\\ 0.70\end{tabular} & \multicolumn{1}{c|}{70.86} & \begin{tabular}[c]{@{}c@{}}0.57/\\ 0.57\end{tabular} & \multicolumn{1}{c|}{56.44} & \begin{tabular}[c]{@{}c@{}}0.49/\\ 0.50\end{tabular} & 49.25 \\ \midrule
HIPT\cite{chen2022scaling} & \begin{tabular}[c]{@{}c@{}}0.41/\\ 0.37\end{tabular} & \multicolumn{1}{c|}{37.94} & \begin{tabular}[c]{@{}c@{}}0.70/\\ 0.74\end{tabular} & \multicolumn{1}{c|}{72.45} & \begin{tabular}[c]{@{}c@{}}0.59/\\ 0.58\end{tabular} & \multicolumn{1}{c|}{57.66} & \begin{tabular}[c]{@{}c@{}}0.41/\\ 0.40\end{tabular} & 39.50 \\ \midrule
KAT\cite{zheng2023kernel} & \begin{tabular}[c]{@{}c@{}}0.37/\\ 0.39\end{tabular} & \multicolumn{1}{c|}{38.30} & \begin{tabular}[c]{@{}c@{}}0.73/\\ 0.72\end{tabular} & \multicolumn{1}{c|}{73.89} & \begin{tabular}[c]{@{}c@{}}0.58/\\ 0.58\end{tabular} & \multicolumn{1}{c|}{58.11} & \begin{tabular}[c]{@{}c@{}}0.53/\\ 0.53\end{tabular} & 53.25 \\ \midrule
UVCGAN\cite{torbunov2023uvcgan} & \begin{tabular}[c]{@{}c@{}}0.41/\\ 0.37\end{tabular} & \multicolumn{1}{c|}{39.87} & \begin{tabular}[c]{@{}c@{}}0.75/\\ 0.75\end{tabular} & \multicolumn{1}{c|}{75.90} & \begin{tabular}[c]{@{}c@{}}0.65/\\ 0.65\end{tabular} & \multicolumn{1}{c|}{65.22} & \begin{tabular}[c]{@{}c@{}}0.47/\\ 0.46\end{tabular} & 46.87 \\ \midrule
DiffAE\cite{preechakul2022diffusion} & \begin{tabular}[c]{@{}c@{}}0.46/\\ 0.46\end{tabular} & \multicolumn{1}{c|}{45.38} & \begin{tabular}[c]{@{}c@{}}0.80/\\ 0.82\end{tabular} & \multicolumn{1}{c|}{80.26} & \begin{tabular}[c]{@{}c@{}}0.70/\\ 0.71\end{tabular} & \multicolumn{1}{c|}{70.33} & \begin{tabular}[c]{@{}c@{}}0.58/\\ 0.59\end{tabular} & 58.37 \\ \midrule
PosDiffAE & \begin{tabular}[c]{@{}c@{}}\textbf{0.47}/\\ \textbf{0.49}\end{tabular} & \multicolumn{1}{c|}{\textbf{49.14}} & \begin{tabular}[c]{@{}c@{}}\textbf{0.81}/\\ \textbf{0.83}\end{tabular} & \multicolumn{1}{c|}{\textbf{81.30}} & \begin{tabular}[c]{@{}c@{}}\textbf{0.75}/\\ \textbf{0.76}\end{tabular} & \multicolumn{1}{c|}{\textbf{75.88}} & \begin{tabular}[c]{@{}c@{}}\textbf{0.69}/\\ \textbf{0.70}\end{tabular} & \textbf{70.12} \\ \bottomrule
\end{tabular}%
}
\label{table:cross_data}
\end{table}
%%%%%%%%%%%%%%%%%%%%%%%%%%%%%%%%%%%%%%%%%%

\subsection{Classification Performance In Cross-Data Setup / Test Datasets}

The classification results for the Cross-Data Setup are presented in Table \ref{table:cross_data}, showcasing the average of precision, recall, and accuracy across various classes. The table indicates that our model, $PosDiffAE$, outperforms other baselines in all the data setups across all metrics. Consistent with the nature of performance observed in Table \ref{table:Main_classif_table}, the diffusion set of models performs better in the Cross-Data Setup compared to the models with a transformer backbone. When the inference is carried out in the $15$ pcw Allen brain, there is an overall reduction in the performance metrics when compared to the scores of the $19$ pcw brain. This can be primarily due to the reason that a $15$ pcw brain tends to be less developed as compared to $19$ and $21$ pcw brains, which might lead to a lack of discriminatory properties in younger brains. The superior performance of metrics for the $19$ pcw brain suggests that classification remains unbiased to the cellular characteristics of a specific brain from a particular data repository. Instead, performance appears to rely on the region-wise discriminative properties available within the dataset, and a lack of such properties will lead to performance degradation. While inferring on the $23$ pcw and $34$-year-old adult brains, there is a decrement in performance relative to the $19$ pcw brains. 
In the case of $23$ pcw, this decrement can be attributed to the sagittal acquisition of the images, unlike coronal acquisitions in the training set. 
In the case of the $34$-year-old adult brain, the textural variability within the image patches reduces, and the region demarcations primarily stem from positional variation. This positional dependency helps our model, $PosDiffAE$, to have a notable performance improvement as compared to other models. However, lack of textural variability leads to lesser discrimination and performance drop compared to $19$, $21$, and $23$ pcw brains.
Furthermore, Figure \ref{fig:FB62} highlights qualitatively that our models can perform reasonably well for the $19$ pcw brain, although the data is acquired from a completely different acquisition setup. Similar to the $21$ pcw Allen dataset, regions that are correlated have a chance of misclassification.

\subsection{Performance of Cross-Data Setup In Artifact Settings}

The classification results in Figure \ref{fig:cross_data_artif} (c) indicate that our model, $PosDiffAE$, generalizes well to the $23$ pcw brain, despite being acquired in a different plane from the training data. Our restoration model, $PosDiffAE+Restore$, improves accuracy for tear and JPEG artifacts. The nature of classification performance for transformer-based models follows a similar nature to the $21$ pcw in Table \ref{table:Atifcat_classif_table}. The transformers trained with spatial masking (SwinMAE, KAT, and UVCGAN) strategies perform relatively better for the tear artifact and have lower accuracies for JPEG compression. 
For the $34$-year-old adult brain (Figure \ref{fig:cross_data_artif} (f)), trends mirror the $23$ pcw brain, though there is an overall decline in the accuracies due to lesser variability in image patches.
The regression results, as given in Figure \ref{fig:cross_data_artif} (a, b, d, and e), highlight that our model performs better across all \textit{Original Sets}. 
For the \textit{Tear Artifact} sets, diffusion models produce higher errors than spatial masking-based transformers. However, our restoration technique achieves lower errors for these sets by retrieving contextually meaningful images. Among the spatial masking-based transformers, JPEG compression has higher errors. While MAE uses spatial masking, its performance differs from other masked transformers as it does not process spatial information for the masked areas.

%%%%%%%%%%%%%%%%%%%%%%%%%%%%%%%%%%%%%
\begin{figure*}[t]
  \centering
  \includegraphics[width=1
\linewidth]{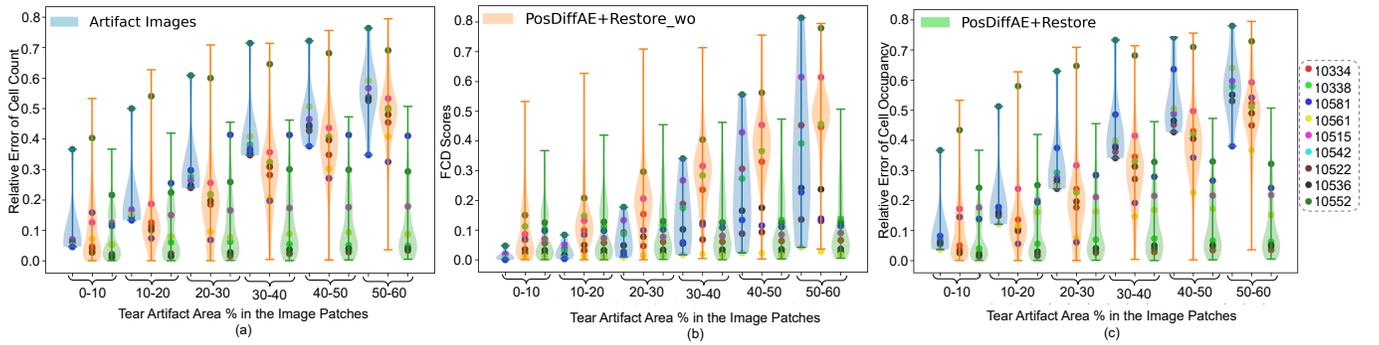}
\caption{The relative error of cell properties vs the proportion of tear artifact affected regions is plotted. The errors evaluated between the Original dataset and (i) the Tear Artifact set, (ii) the PosDiffAE+Restore\_wo, and (iii) the PosDiffAE+Restore restoration sets are highlighted in  (a) and (c). Similarly, plot (b) showcases the FCD scores within the sets. For all the plots, the errors and scores of different regions are highlighted with multi-coloured dots, and the inter-class mean and variance of them are also shown.}
\label{fig:tear_artifact_plot}
\end{figure*}
%%%%%%%%%%%%%%%%%%%%%%%%%%%%%%%%%%%%%

%%%%%%%%%%%%%%%%%%%%%%%%%%%%%%%%%%%%%%
\begin{figure}[!b]
  \centering
  \centerline{\includegraphics[width=1\linewidth]{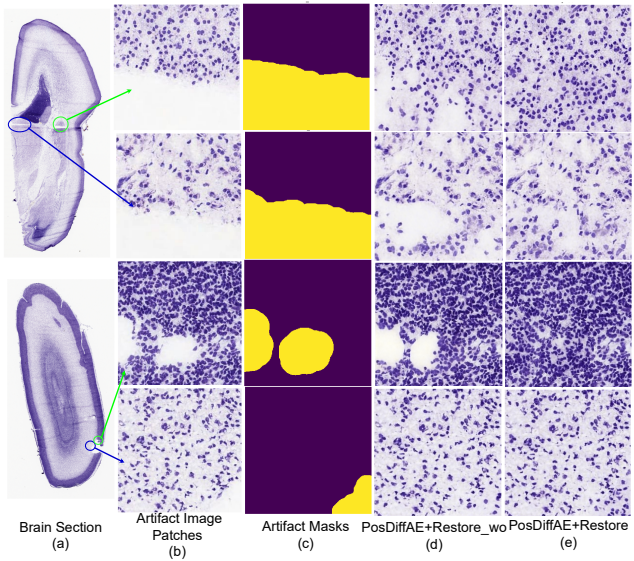}}
\caption{Reconstructed images from the Tear Artifact set. From left to right, (a) Brain sections having artifacts, (b) Image patches from the Tear Artifact set, (c) Artifact masks, reconstructions from (d) PosDIffAE+Restore\_wo and (e) PosDIffAE+Restore.}
\label{fig:Tear_artifact_image}
\end{figure}
%%%%%%%%%%%%%%%%%%%%%%%%%%%%%%%%%%%%%

\subsection{Analysis Of Artifact Restorations }
 \textit{What is the effect on Cellular Properties due to Tear Artifact and restoration?} The plots in Figure \ref{fig:tear_artifact_plot} show the error between the cellular properties of the Original dataset and various artifact/ restoration sets comprising of the (i) \textit{Tear Artifact} set, (ii) $PosDiffAE+Restore\_wo$, and (iii) $PosDiffAE+Restore$ restored sets. The cellular properties considered here are (a) cell count of image patches, (b) FCD score between datasets, and (c) cell occupancy of image patches. All the properties shown in Figure \ref{fig:tear_artifact_plot} are evaluated for the nine regions of the Allen dataset separately. From the cell count and cell occupancy plots (Figure \ref{fig:tear_artifact_plot} a, c), it is evident that the error of the \textit{Tear Artifact} set increases steeply compared to the other two restoration sets. Although the error of the $PosDiffAE+Restore\_wo$ set is lower than the \textit{Tear Artifact} set, it deviates significantly from the $PosDiffAE+Restore$ set, especially for higher artifact-affected images. For the \textit{Tear Artifact} set, as the artifact-affected area (\%) increases, image patches from region CGE ($10552$) show higher error rates due to higher cell density, while regions with lower cell density, like ic ($10581$), have lower error rates. For the $PosDiffAE+Restore$ set, both low and high-density regions exhibit higher error rates since generation for the extreme cases is typically more challenging. The mean errors of various regions for the $PosDiffAE+Restore$ set lie below the $0.1$ error value, while the other sets have higher values. The plot of the FCD scores (Figure \ref{fig:tear_artifact_plot} b) shares similar trends compared to the other plots, but the region-wise FCD error is not affected for the low-density regions in the \textit{Tear Artifact} set.

The nature of errors for the $PosDiffAE+Restore\_wo$ set deviates from the $PosDiffAE+Restore$ set since, without the neighborhood guidance, the restoration occurs in an uncontrolled manner, generating images from anywhere within the entire distribution. 
%
% the restored image can not move towards the distribution of the specific region the restoration is intended to be. In the absence of guidance, the restoration happens in an uncontrolled manner, generating images from anywhere within the entire distribution. 
This is because the tear artifact-affected regions may resemble some other region of the brain, which has a mix of cellular and whitish-appearing regions. As shown in Figure \ref{fig:Tear_artifact_image}, the restored (Figure \ref{fig:Tear_artifact_image} e) images have more contextually relevant region-level information when the neighborhood guidance is available.
%%%%%%%%%%%%%%%%%%%%%%%%%%%%%%%%%%%%%%%

% \usepackage{booktabs}
% \usepackage{graphicx}
\renewcommand{\arraystretch}{0.8} 
\begin{table*}[t]
\caption{Quantitative evaluation for JPEG Restoration with different models and various levels of compression ($21$ pcw validation dataset) and $^*$ denotes statistical significance ($p<0.01$)}.

\centering
\resizebox{1.7\columnwidth}{!}{%
\begin{tabular}{@{}cccccccccc@{}}
\toprule
Method            & \multicolumn{3}{c}{PSNR ($\uparrow$)}                   & \multicolumn{3}{c}{SSIM ($\uparrow$)}                & \multicolumn{3}{c}{FCD ($\downarrow$)}  \\ \midrule
                  & QF 5         & QF 10        & QF 15        & QF 5        & QF 10       & QF 15       & QF 5   & QF 10  & QF 15  \\ \midrule
JPEG Artifact     & 22.94$\pm$3.36 & 25.16$\pm$3.21 & 26.50$\pm$3.46 & 0.74$\pm$0.09 & 0.81$\pm$0.05 & 0.85$\pm$0.04 & 0.2630 & 0.1468 & 0.1036 \\ \midrule
FBCNN\cite{jiang2021towards}            & 25.33$\pm$3.01 & 26.09$\pm$2.99 & 26.92$\pm$3.23 & \textbf{0.89$\pm$0.04} & \textbf{0.91$\pm$0.03} & \textbf{0.93$\pm$0.02} & 0.1429 & 0.1186 & 0.0896 \\ \midrule
DDRM\cite{kawar2022denoising}               & 22.96$\pm$3.72 & 25.01$\pm$3.82 & 25.86$\pm$4.00 & 0.76$\pm$0.10 & 0.83$\pm$0.06 & 0.86$\pm$0.05 & 0.1961 & 0.1000 & 0.0734 \\ \midrule
PosDiffAE+Restore\_wo & 26.04$\pm$3.67 & 26.42$\pm$3.78 & 26.55$\pm$3.75 & 0.86$\pm$0.05 & 0.87$\pm$0.04 & 0.87$\pm$0.04 & 0.0589 & 0.0560 & 0.0534 \\ \midrule
PosDiffAE+Restore & \textbf{26.67$^*$$\pm$3.58} & \textbf{27.15$^*$$\pm$3.69} & \textbf{27.31$^*$$\pm$3.65} & \textbf{0.89$^*$$\pm$0.04} & 0.90$^*$$\pm$0.04 & 0.91$^*$$\pm$0.03 & \textbf{0.0422$^*$} & \textbf{0.0394$^*$} & \textbf{0.0368$^*$} \\ \bottomrule
\end{tabular}%
}
\label{table:JPEG_artifact}
\end{table*}
%%%%%%%%%%%%%%%%%%%%%%%%%%%%%%%%%%%%%

%%%%%%%%%%%%%%%%%%%%%%%%%%%%%%%%%%%%%%
\begin{figure}[]
  \centering
  \centerline{\includegraphics[width=1\linewidth]{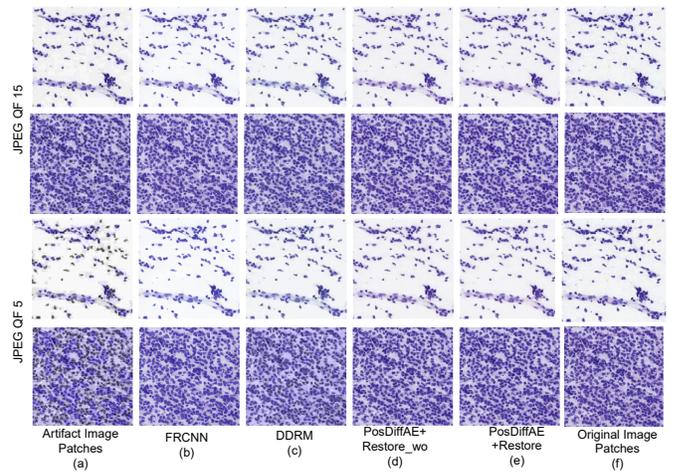}}
\caption{Restored images for two different levels of JPEG compression (top two rows QF 15 and bottom two rows QF 5). From left to right, columns represent (a) Artifact Images, model outputs, (b) FRCNN, (c) DDRM, (d) PosDiffAE+Restore\_wo, (e) PosDiffAE+Restore, and (f) Original ground truth images. Notably, the PosDiffAE set of models can better preserve the textural content of the images, resulting in a better FCD score.  }
\label{fig:JPEG_artifact_image}
\end{figure}
%%%%%%%%%%%%%%%%%%%%%%%%%%%%%%%%%%%%%%%%
% FBCNN \cite{jiang2021towards}, DDRM \cite{kawar2022denoising}
% Table \ref{table:JPEG_artifact} Figure \ref{fig:Tear_artifact_image}
\subsubsection{Analysis of JPEG Restoration} \label{subsubsec:Analysis_JPEG_Restoration} The Table \ref{table:JPEG_artifact} highlights that our model $PosDiffAE+Restore$ performs better than the others in terms of PSNR and FCD. 
As compared to the diffusion-based model DDRM \cite{kawar2022denoising}, our model has higher PSNR and lower FCD, implying that our adaptive noising strategy could preserve the faithfulness and realism of the generated images.
The performance of the ablated version of our model ($PosDiffAE+Restore\_wo$) shows that guidance is essential for better faithfulness and realism of the generated images. The CNN-based model, FBCNN \cite{jiang2021towards}, has better SSIM for a lower compression rate since this model tends to blur out the original textural appearance of the images, which elevates the structural appearance of the cells. However, this results in a higher drop in FCD, indicating reduced realism.
These claims are further supported by the images in Figure \ref{fig:JPEG_artifact_image}. These results suggest that diffusion models effectively restore artifact images with reasonable realism and fidelity.
% , even without prior knowledge of the nature of the artifact.

\subsubsection{Analysis of Black-dot artifact restoration} 
\begin{figure}[]
  \centering
  \includegraphics[width=1
\linewidth]{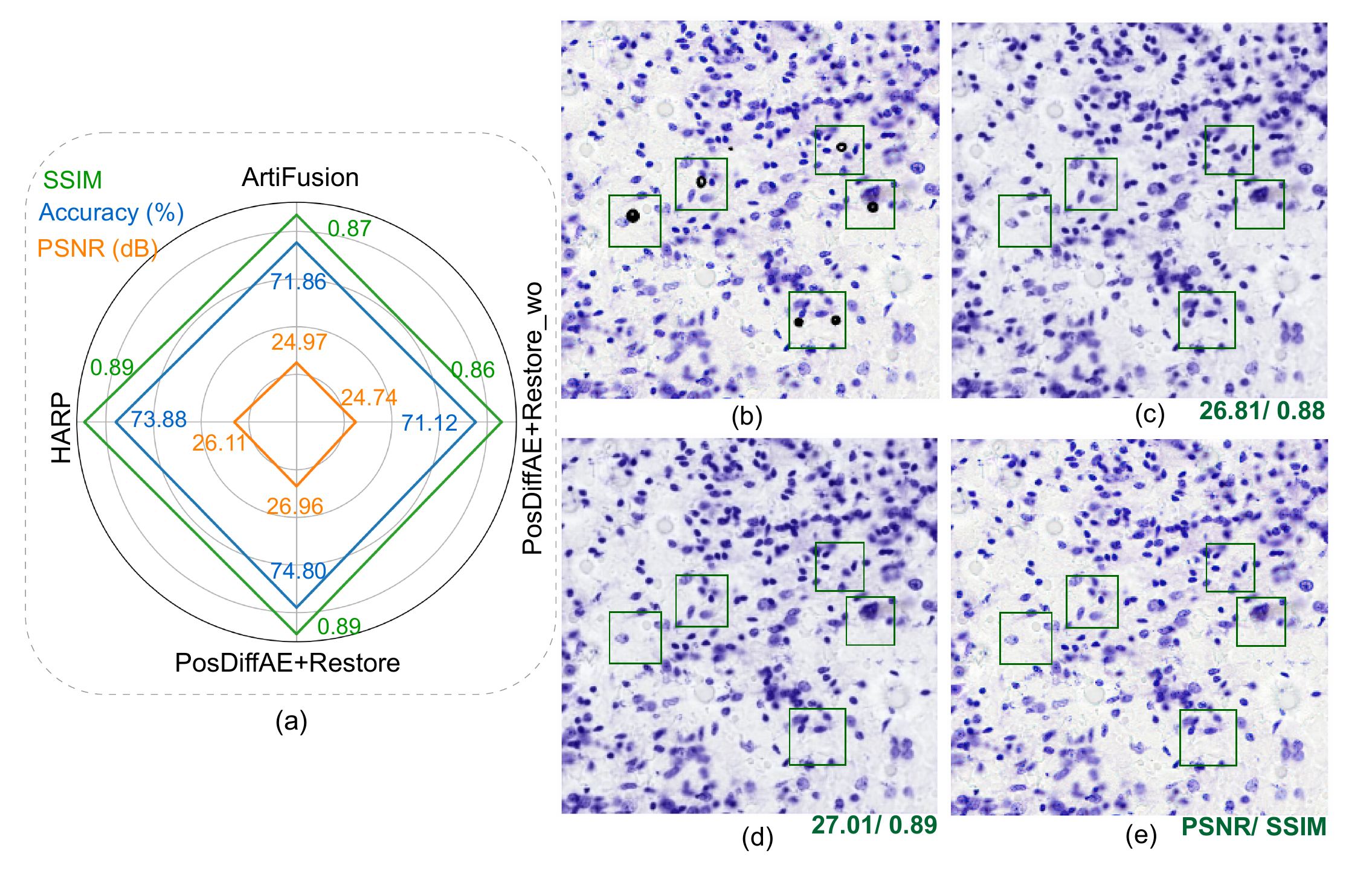}
\caption{The figure represents various metrics and images after reconstruction of the Black-dot artifact set ($23$ pcw from the test dataset). From left to right, (a) represents the radar chart of different reconstruction results for various methods, (b) black-dot artifact image, reconstructed images from (c) HARP, (d) PosDiffAE+Restore, and (e) the original image. The green boxes highlight the regions of artifact. } 
\label{fig:black_dot_artifact}
\end{figure}

The Figure \ref{fig:black_dot_artifact} (a) highlights that $PosDiffAE$ performs relatively better than the other diffusion-based baselines in terms of reconstruction metrics (PSNR and SSIM) and classification accuracy of the reconstructed images. The neighborhood awareness-based reconstruction strategy has enabled our model to generate better contextually aware tissue regions. Among the comparison models, the HARP \cite{fuchs2024harp} model is able to perform better since the method integrates specialized techniques to extract the regions of the artifact. ArtiFusion \cite{he2023artifact} performs better than PosDiffAE+Restore\_wo since the former has a Swin-transformer incorporated into the diffusion model. The reconstructed images of HARP and our model are shown in Figure \ref{fig:black_dot_artifact} (c) and (d), respectively. The reconstructions from our model more closely resemble the original, whereas HARP primarily generates cellular structures in areas with black dots.
  
% \cite{he2023artifact} and HARP \cite{fuchs2024harp}
% Specifically, the difference in textural appearance between the FRCNN and the $PosDiffAE$ set of models is evident in Figure \ref{fig:JPEG_artifact_image} (b) and (d, e). 
%%%%%%%%%%%%%%%%%%%%%%%%%%%%%%%%%%%%%
%%%%%%%%%%%%%%%%%%%%%%%%%%%%%%%%%%%%%

\section{Conclusion}
\label{sec:conclusion}

In this paper, we proposed a Diffusion Auto-encoding model for processing histological images of neuronal tissue. We demonstrate the applicability of the model in region classification, spatial information regression, and artifact restoration by leveraging the meaningful representational space of the Diffusion AE. We have showcased the performance improvement of our method in discrimination, regression, and restoration tasks, highlighting the discriminative and guiding properties of the latent vectors encoded by diffusion models. 
\textit{Limitations and Future Scope}: While we have focused on the expansive regions of the brain evident from the Nissl stain, extending to histochemical and antibody stains can potentially extend to regions specific to these stains. Furthermore, topological information can be integrated to differentiate regions where texture and positional information are inadequate; this can yield better outcomes for discriminating regions like Ca and Pu. Additionally, our current scope of work focuses on restoring tear, JPEG, and black-dot artifacts, which can be expanded to more artifacts, like blurring due to water bubbles. Extension to multiple artifacts can be achieved by integrating an artifact localization algorithm prior to restoration. Finally, our work demonstrates diffusion models as potential representation learners for histological images, capable of driving multiple downstream tasks for histological image analyses in an unsupervised manner.
%

% Our approach currently handles a limited number of brain regions and operates on a fixed patch size, focusing on cellular-level information. There is a reliance of 
% our approach on a specific staining type and positional data from co-registered brain sections. Future work can focus on relaxing some of the dependencies through methodological improvements.
% The future scope further includes larger sizes of the training set, which can be expanded to yield better outcomes and detect more regions of the brain automatically. The restoration capability of our method can be expanded to more artifacts, like blurring due to water bubbles. This can be achieved by integrating an artifact detection algorithm prior to restoration. 
% Our work demonstrates the capability of Diffusion models as potential feature extractors, highlighting the utilizability of these features for downstream histological image analyses.

% of the work
% \bibliographystyle{elsarticle-num} 
% \bibliography{PosDiffAE}
% \input{article.bbl}




\printbibliography

\end{document}